\pdfoutput=1
\documentclass{article}

\usepackage{times}
\usepackage[margin=1in]{geometry}
\usepackage{natbib}

\usepackage[utf8]{inputenc}
\usepackage[T1]{fontenc}
\usepackage{url}
\usepackage{nicefrac}
\usepackage{microtype}

\usepackage{amsmath,amssymb,amsfonts,amsthm}
\usepackage{algorithmic}
\usepackage{algorithm}
\newtheorem{definition}{Definition}
\usepackage{graphicx}
\usepackage{textcomp}
\usepackage{xcolor}
\usepackage{booktabs}
\usepackage{multirow}
\usepackage{subcaption}
\usepackage{placeins}
\usepackage{xspace}
\usepackage{hyperref}
\usepackage{cleveref}
\usepackage{tikz}
\usetikzlibrary{arrows.meta, positioning, shapes.geometric, fit, backgrounds, calc}

\usepackage{amsmath,amsfonts,bm}

\def\eqref#1{equation~\ref{#1}}

\def\1{\bm{1}}

\DeclareMathAlphabet{\mathsfit}{\encodingdefault}{\sfdefault}{m}{sl}
\SetMathAlphabet{\mathsfit}{bold}{\encodingdefault}{\sfdefault}{bx}{n}

\newcommand{\eg}{e.g.\@\xspace}

\newcommand{\Gcal}{\mathcal{G}}
\newcommand{\Kcal}{\mathcal{K}}

\newcommand{\Lcal}{\mathcal{L}}
\newcommand{\Acal}{\mathcal{A}}
\newcommand{\Dcal}{\mathcal{D}}
\newcommand{\Ein}{E_{\mathrm{in}}}
\newcommand{\Edec}{E_{\mathrm{decided}}}
\newcommand{\Eeval}{E_{\mathrm{evaluated}}}
\newcommand{\Eret}{E_{\mathrm{retrieved}}}
\newcommand{\pcov}{P_{\mathrm{cov}}}
\newcommand{\dtau}{D_{\tau}}
\newcommand{\atau}{A_{\tau}}

\title{ROZA Graphs: Self-Improving Near-Deterministic RAG through Evidence-Centric Feedback}

\author{Matthew Penaroza}
\date{\today}

\begin{document}

\maketitle

\begin{abstract}
Language model agents reason from scratch on every query, discarding their chain of thought after each run. This produces lower accuracy and high variance. We introduce \emph{reasoning graphs}, a graph structure that persists per-evidence chain of thought as structured edges. Unlike prior memory mechanisms that retrieve distilled strategies by query similarity, reasoning graphs enable \emph{evidence-centric feedback}: each evidence item's full verdict history is retrieved by backward traversal across all prior runs. We further introduce \emph{retrieval graphs}, a complementary structure that closes the retrieval loop by feeding a pipeline planner that tightens the candidate funnel over successive runs. Together, both graphs form a ROZA graph, a self-improving feedback loop whose gains are gated on two reuse axes: accuracy on gold-passage reuse via the reasoning graph, efficiency on candidate-pool overlap via the retrieval graph. The base model remains frozen; all gains come from context engineering via graph traversal. We evaluate on MuSiQue, HotpotQA, and 2WikiMultiHopQA. The accuracy mechanism produces a clean dose-response on MuSiQue: at zero evidence-profile coverage the system is by design identical to Vanilla RAG and accuracy is tied (\(\sim\)54\% on both); the gap then grows monotonically, reaching $+10.6$pp at 50\%+ coverage on the same questions (47\% error reduction, $p<0.0001$). The per-question Spearman rank correlation between coverage and the paired accuracy delta is $\rho{=}{+}0.144$ (95\% CI $[+0.087,+0.201]$, one-sided $p<10^{-6}$, $n{=}1{,}100$), confirming the dose-response without binning or stratification. The mechanism scales with reasoning depth: 4-hop accuracy improves by $+11.0$pp ($p=0.0001$). Cross-dataset, the cluster-level accuracy gain is predicted by gold-passage reuse density ($r=0.604$, $p=0.001$, $n=26$ clusters), and 2WikiMultiHopQA returns a pre-registered null at low gold-reuse ($+0.3$pp, $p=0.77$). In high-reuse deployments where candidate-pool overlap is dense, the retrieval graph delivers Pareto dominance: highest or statistically tied for highest accuracy, 46\% lower cost, and 46\% lower latency. On a paired determinism probe set ($N{=}73$, $K{=}10$ runs per query, two model families, three sampling temperatures), evidence-centric feedback substantially increases per-passage decision consistency across runs: average consistency rises by $+8$ to $+13$pp on a fixed 20-passage context (matched to the baseline; only the evidence profile changes), and by $+12$ to $+21$pp when the retrieval graph also prunes (all $p<0.005$).
\end{abstract}

\section{Introduction}
\label{sec:intro}

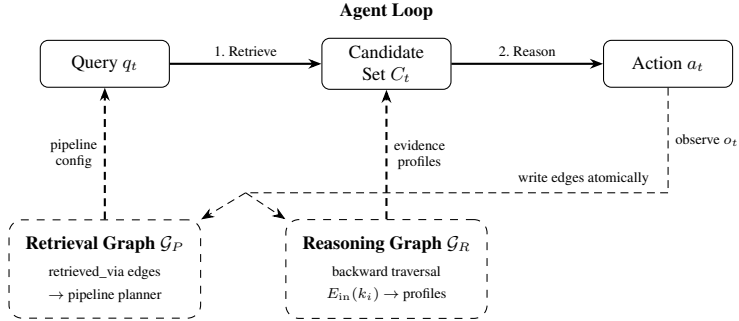
\begin{figure}[t]
    \centering
    \begin{tikzpicture}[scale=0.78, transform shape,
        box/.style={draw, rounded corners=3pt, minimum height=0.8cm, minimum width=2.2cm, align=center, font=\small},
        graphbox/.style={draw, dashed, rounded corners=5pt, inner sep=8pt, align=center, font=\small},
        arr/.style={-{Stealth[length=5pt]}, thick},
        lbl/.style={font=\scriptsize, align=center},
    ]
    \node[box] (query) at (0,0) {Query $q_t$};
    \node[box] (cand) at (4.8,0) {Candidate\\Set $C_t$};
    \node[box] (action) at (9.6,0) {Action $a_t$};

    \draw[arr] (query) -- node[above, lbl] {1.~Retrieve} (cand);
    \draw[arr] (cand) -- node[above, lbl] {2.~Reason} (action);
    \node[font=\small\bfseries] at (4.8, 0.9) {Agent Loop};

    \node[graphbox] (pg) at (0, -3.5) {\textbf{Retrieval Graph} $\Gcal_P$\\[2pt]{\scriptsize retrieved\_via edges}\\{\scriptsize $\to$ pipeline planner}};
    \node[graphbox] (rg) at (4.8, -3.5) {\textbf{Reasoning Graph} $\Gcal_R$\\[2pt]{\scriptsize backward traversal}\\{\scriptsize $\Ein(k_i)$ $\to$ profiles}};

    \draw[arr, densely dashed] (pg.north) -- node[left, lbl] {pipeline\\config} (query.south);
    \draw[arr, densely dashed] (rg.north) -- node[right, lbl] {evidence\\profiles} (cand.south);

    \coordinate (mid) at (2.4, -2.2);
    \coordinate (drop) at (9.6, -2.2);
    \draw[densely dashed] (action.south) -- node[right, lbl] {observe $o_t$} (drop);
    \draw[densely dashed] (drop) -- node[above, lbl, pos=0.2] {write edges atomically} (mid);
    \draw[densely dashed, -{Stealth[length=5pt]}] (mid) -- (pg.north east);
    \draw[densely dashed, -{Stealth[length=5pt]}] (mid) -- (rg.north west);
    \end{tikzpicture}
    \caption{Architecture. The agent loop (retrieve, reason, act) is augmented with two feedback graphs: the \emph{reasoning graph} $\Gcal_R$ feeds evidence profiles via item-inward traversal of past evaluations; the \emph{retrieval graph} $\Gcal_P$ feeds a pipeline planner that excludes consistently-rejected items. Both update atomically after each run. Evidence-centric feedback starts from the specific items in $C_t$, not from query-embedding similarity.}
    \label{fig:architecture}
\end{figure}

Every language-model agent follows the same loop (retrieve, reason, act~\citep{yao2023react, sumers2023cognitive}) and discards its chain of thought after each step, leaving accuracy lower than it needs to be (the agent re-derives evaluations on recurring evidence) and variance high (the same query type succeeds or fails unpredictably)~\citep{agentconsistency2025}. Prior memory mechanisms (Reflexion~\citep{shinn2023reflexion}, ReasoningBank~\citep{ouyang2025reasoningbank}) are \emph{query-centric}: they retrieve distilled strategies by query similarity, with no direct connection to the specific evidence items currently in front of the agent. We instead propose \emph{evidence-centric feedback}: \emph{reasoning graphs} persist the per-evidence chain of thought as graph edges with bidirectional traversal, so the system can ask, for every candidate item, ``what does it already know about \emph{this specific item}?'', traversing all incoming evaluation edges across prior runs to construct an \emph{evidence profile} of typical verdicts, reasoning patterns from correct decisions, and a reliability signal (Figure~\ref{fig:architecture}).

Persisting evidence-level reasoning solves post-retrieval adjudication but retrieval itself stays static; we therefore introduce a complementary \emph{retrieval graph} that feeds a pipeline planner to prune consistently-rejected candidates. The two graphs operate on different reuse axes: the retrieval graph drives \emph{efficiency} through candidate-pool overlap (prunes whether-or-not-gold), and the reasoning graph drives \emph{accuracy} through gold-passage recurrence (surfaces prior verdicts on the answer-critical items). The axes are dissociable: 2WikiMultiHopQA has near-MuSiQue Jaccard overlap (0.499) but near-zero gold-passage reuse (0.025), and the system returns a pre-registered accuracy null ($+0.3$pp) while the retrieval-graph efficiency story is preserved.

\paragraph{Contributions.} (i) \textbf{Reasoning graphs} with evidence-centric bidirectional traversal (Section~\ref{sec:reasoning_graphs}); (ii) \textbf{retrieval graphs} feeding a pipeline planner that prunes consistently-rejected candidates (Section~\ref{sec:retrieval_graphs}); (iii) \textbf{ROZA graphs}, a self-improving feedback architecture that compounds both graphs into accuracy convergence and verdict-level variance collapse without retraining (Section~\ref{sec:feedback_loop}); (iv) \textbf{empirical evaluation} on MuSiQue, HotpotQA, and 2WikiMultiHopQA (47\% error reduction at high coverage, verdict-vector identity rate (VVIR; Table~\ref{tab:vvir}), Pareto-dominant efficiency in high-reuse, with component ablations) (Section~\ref{sec:experiments}).

\section{Related Work}
\label{sec:related}

\paragraph{Memory and self-improvement for LLM agents.}
Reflexion~\citep{shinn2023reflexion} stores verbal self-reflections in a flat text buffer; Voyager~\citep{wang2023voyager} maintains a skill library; MEM1~\citep{zhou2026mem1} learns a compact memory state via RL. Most relevant is ReasoningBank~\citep{ouyang2025reasoningbank}, which distills strategies from agent experiences as flat triples retrieved by query-embedding similarity. Our work differs in both mechanism and capability: ReasoningBank is \emph{query-centric}: strategies are distilled away from their source evidence and cannot answer ``how has evidence item $k_i$ been evaluated across all past runs?'' Reasoning graphs are \emph{evidence-centric}: backward traversal from each item inward across all prior evaluations is the load-bearing graph operation. Conceptual predecessors include case-based reasoning~\citep{aamodt1994case, kolodner1993case}, the CoALA agent-memory taxonomy~\citep{sumers2023cognitive}, episodic memory in Generative Agents~\citep{park2023generative}, experience replay for gradient-based policy updates~\citep{lin1992self, mnih2015human}, and stronger-to-weaker memory transfer~\citep{preconditions2025memory}; all store flat records and lack the per-evidence backward traversal that reasoning graphs require.

\paragraph{Retrieval and graph-structured RAG.}
RAG~\citep{lewis2020retrieval, gao2023retrieval} retrieves passages to augment LM inputs. Self-RAG~\citep{asai2023self} optimizes retrieval \emph{within} a run via learned reflection tokens; reasoning graphs optimize \emph{across} runs via persistent graph structure, orthogonal and composable. GraphRAG and follow-ups~\citep{edge2024local, linearrag2026, whentoGraphRAG2026, peng2024graph} structure the \emph{source data} as graphs to improve retrieval; we structure \emph{reasoning about} the source data as graphs. GAM-RAG~\citep{gamrag2026} and EvalAct~\citep{evalact2026} share the cross-query adaptation motivation but operate on retrieval-statistical mechanisms (Kalman gain rule, process-calibrated rewards); they are orthogonal to evidence-centric reasoning and could be composed with our reasoning graph.

\paragraph{Behavioral consistency and decision stability.}
Recent empirical studies quantify behavioral inconsistency in LM agents~\citep{agentconsistency2025}, finding 2.0--4.2 distinct action sequences per 10 runs on identical HotpotQA inputs and a 32--55pp accuracy gap between consistent and inconsistent tasks. Follow-up work~\citep{consistencyamplifies2025} shows consistency can amplify both correct and incorrect interpretations; 71\% of failures in one model stem from ``consistent wrong interpretation.'' Our evidence profiles address this through the correct-outcome filter (Section~\ref{sec:reasoning_graphs}): the agent receives history from decisions verified correct, blocking amplification of incorrect reasoning patterns.

\section{Method}
\label{sec:method}

\subsection{Problem Formalization}
\label{sec:formalization}

An agent over knowledge base $\Kcal{=}\{k_1,\ldots,k_n\}$ executes three steps at each timestep~$t$ given query $q_t$: \textbf{Retrieve} $R(q_t)\to C_t\subseteq\Kcal$ (candidate set of $m$ items); \textbf{Reason} $\Lcal(q_t,C_t)\to(a_t,\theta_t)$ where $a_t$ is the answer and $\theta_t{=}\{(k_i,v_i,r_i,\delta_i)\}_{k_i\in C_t}$ is the structured chain of thought, with verdict $v_i\in\{\texttt{used},\texttt{rejected}\}$, natural-language reason $r_i$, and confidence delta $\delta_i\in[-1,1]$; \textbf{Observe} outcome $o_t\in\{\texttt{correct},\texttt{incorrect}\}$. Task accuracy is $\atau(n){=}\tfrac{1}{n}\sum_{t=1}^n\mathbb{1}[o_t{=}\texttt{correct}]$ for type~$\tau$. Standard agents have two structural limits: \emph{cold-start reasoning} ($\theta_t$ independent of $\theta_{1\ldots t-1}$) and \emph{static retrieval} ($R$ fixed across timesteps); reasoning graphs address the first, retrieval graphs the second.

\subsection{Reasoning Graphs}
\label{sec:reasoning_graphs}

\begin{definition}[Reasoning Graph]
A reasoning graph $\Gcal_R = (V, \Edec \cup \Eeval)$ is a directed labeled property graph where:
\begin{itemize}
    \item $V = \Acal \cup \Dcal \cup \Kcal$ comprises agent nodes, decision nodes, and evidence item nodes.
    \item $\Edec = \{(a, d, \phi) \mid a \in \Acal, d \in \Dcal\}$ are \emph{decided} edges, where $\phi = (\text{confidence}, \text{timestamp}, \text{outcome})$. The outcome field is populated after a downstream signal is received.
    \item $\Eeval = \{(a, k_i, \psi) \mid a \in \Acal, k_i \in \Kcal\}$ are \emph{evaluated} edges, where $\psi = (\text{step}, \text{verdict}, \text{reason}, \text{confidence\_delta}, \text{action\_ref})$.
\end{itemize}
\end{definition}

\paragraph{Write protocol and evidence-centric context injection.} After the LM produces $(a_t,\theta_t)$ via structured output, the system atomically writes one decided edge $(a,d_t,\phi_t)$ and $|C_t|$ evaluated edges $\{(a,k_i,\psi_i)\}$; the outcome field on $\phi_t$ is filled in when the downstream signal arrives. For each new query, evidence profiles are constructed by (i) traversing incoming evaluated edges $\Ein(k_i){=}\{e\in\Eeval\mid e{=}(*,k_i,*)\}$ for every $k_i\in C_t$, (ii) keeping only edges whose decided edge has $\text{outcome}{=}\texttt{correct}$ (the \emph{correct-outcome filter}), (iii) aggregating verdicts/reasons and computing the reliability score
\begin{equation}
R(k_i,\tau) \;=\; \frac{|\{e\in\Ein(k_i):e.\text{verdict}{=}\texttt{used}\wedge e.\text{outcome}{=}\texttt{correct}\}|}{|\Ein(k_i)|},
\end{equation}
and (iv) injecting the profile $\mathcal{P}(k_i)$ alongside the raw evidence. The model thus never reasons from scratch on items it has already evaluated. Backward (item-inward) traversal with outcome-based filtering requires the graph structure: a flat strategy store indexed by query embedding cannot perform it.

\paragraph{Selection policy, properties, and graceful degradation.} Context is bounded by a token budget $B$ and a per-item cap $T_{\max}$ (sample the $N$ most-recent correct-outcome evaluations); items with $|\Ein(k_i)|{=}0$ trigger from-scratch reasoning. Reasoning graphs are auditable (every decision is one traversal away from its full chain of thought), require no retraining, and self-improve as a side effect of normal operation. $R(k_i,\tau)$ doubles as a knowledge-base quality signal: low-$R$ items are consistently misleading, items whose rejection reasons cluster on ``outdated'' are stale, and high-$R$ items are reliable. \emph{Graceful degradation:} at cold start ($|\Ein(k_i)|{=}0$ for all $k_i$) the agent matches the no-memory baseline ($\sim$54\% at 0\% coverage; Sec.~\ref{sec:results}), confirming no infrastructure penalty.

\subsection{Retrieval Graphs}
\label{sec:retrieval_graphs}\label{sec:feedback_loop}

\begin{definition}[Retrieval Graph]
A retrieval graph $\Gcal_P$ augments the reasoning graph with \emph{retrieved\_via} edges: $\Eret = \{(d, q_{\text{source}}, \rho)\}$, where $d \in \Dcal$ is a decision node, $q_{\text{source}}$ is the query or entity that initiated retrieval, and $\rho = (\text{filters}, |C_{\text{pre}}|, |C_{\text{post}}|, \text{top\_relevant}, \text{latency}, \text{token\_cost}, \text{timestamp})$.
\end{definition}

\paragraph{Pipeline planner, query types, and feedback loop.} Given type $\tau$ (dataset labels where available, e.g., HotpotQA bridge vs.\ comparison; otherwise embedding-based $k$-means, MuSiQue $k{=}25$, sensitivity in Appendix~\ref{app:clustering}), the planner traverses $\Gcal_P$ for past retrieved\_via edges of type $\tau$, picks the filter configuration $f^\star{=}\arg\max_f S(f)$ subject to a min-support count where $S(f)$ is per-filter outcome success rate, and applies a cross-graph optimization that excludes evidence items with rejection rate above $R_{\text{thresh}}$. Tightening the candidate funnel cuts both prompt tokens and the number of evaluated edges written per run; in high-reuse settings savings reach $46\%$, while in low-reuse settings traversal overhead can exceed savings (Section~\ref{sec:highreuse}). The two graphs compose into a self-improving loop: retrieval informs \emph{what} to retrieve, the reasoning graph informs \emph{how} to reason about it, and both update atomically. We define decision consistency $\dtau(n)$ as the fraction of items in $C_t$ whose verdict matches the profile majority; as profiles densify, $\dtau(n)$ rises and the variance of $\atau(n)$ collapses; ``deterministic'' here means identical verdicts on identical evidence, not identical tokens. A worked Palme d'Or example, an empirical-convergence argument, and the high-ceiling decomposition appear in Appendix~\ref{sec:enh_method_supp}.

\section{Experiments}
\label{sec:experiments}

\subsection{Setup}
\label{sec:setup}

\paragraph{Models.} Claude Sonnet 4 (primary) and Haiku 4.5 (secondary) via Anthropic API at $T{=}0$ unless noted, base model fixed across configurations within each experiment. A cross-vendor replication on GPT-5-mini (OpenAI, $T{=}1.0$; Appendix~\ref{sec:enh_crossvendor}) confirms generality beyond Anthropic.

\paragraph{Datasets.} Four settings, all using the standard distractor protocol with the gold passage always present (20 passages per question for MuSiQue and High-Reuse, 10 for HotpotQA and 2WikiMultiHopQA): \textbf{MuSiQue}~\citep{trivedi2022musique} 1,100 multi-hop 2--4-step questions in 11 clusters of 100 (primary); \textbf{HotpotQA}~\citep{yang2018hotpotqa} 500 questions in 5 clusters (1 bridge, 4 comparison); \textbf{2WikiMultiHopQA}~\citep{ho2020constructing} 1,000 questions (10 clusters of 100) from the dev distractor split; per-cluster Jaccard candidate-pool overlap 0.499 (near MuSiQue's 0.504) but gold-passage reuse 0.025 (near HotpotQA's 0.030), giving a natural null for the gold-reuse prediction (Sec.~\ref{sec:hop_depth}); and \textbf{MuSiQue High-Reuse} 387 high-Jaccard questions simulating deployment, run on both models. The retrieval graph prunes within these fixed candidate sets; retrieval recall is not varied.

\paragraph{Sequential Cluster Protocol.} We cluster queries into types $\tau$ via $k$-means embeddings (MuSiQue $k{=}25$; High-Reuse 1) or dataset labels (HotpotQA: bridge vs.\ comparison), then process them sequentially per cluster, accumulating reasoning- and retrieval-graph edges across the sequence. We report Acc (token F1 $\geq 0.8$ for MuSiQue, EM for HotpotQA) and mean F1.

\paragraph{Outcome Signal and Determinism Probes.} $o_t{=}\texttt{correct}$ iff token-level F1 $\geq 0.8$, computed automatically against ground truth; the correct-outcome filter (Sec.~\ref{sec:reasoning_graphs}) dilutes noisy labels across many evaluations and robustness is discussed in Sec.~\ref{sec:limitations}. The verdict-determinism experiment uses $N{=}73$ paired probes from MuSiQue High-Reuse (the original 30 probes prospectively expanded with 43 broader-difficulty probes; Appendix~\ref{app:probe_sensitivity}), $K{=}10$ runs each at cp387 across three conditions (Vanilla RAG, Ours-RG, Ours-Full) $\times$ three temperatures $\times$ two models (Haiku 4.5, Sonnet 4).

\subsection{Baselines and Configurations}
\label{sec:baselines}

\begin{itemize}
    \item \textbf{Vanilla RAG}: vector retrieval with no cross-run feedback.
    \item \textbf{Reflexion}~\citep{shinn2023reflexion}: verbal self-reflection stored as flat text, retrieved by recency.
    \item \textbf{ReasoningBank}~\citep{ouyang2025reasoningbank}: we reimplement the core mechanism (distilling strategy triples from each run's chain of thought and retrieving the top-3 most similar strategies by cosine similarity of query embeddings using BAAI/bge-base-en-v1.5) within our agent framework so that all methods share the same retrieval pipeline, prompt structure, and evaluation harness. This shared-framework design ensures that performance differences reflect the feedback mechanism, not confounding implementation choices (different retrievers, prompts, or LLMs).
    \item \textbf{Reasoning graph only} (Ours-RG): evidence-centric injection via reasoning graph, no retrieval graph. This ablation isolates the contribution of evidence profiles.
    \item \textbf{Full system} (Ours-Full): both reasoning and retrieval graphs with evidence-centric injection and pipeline planner.
\end{itemize}

\subsection{Metrics}
\label{sec:metrics}

We report (a) \textbf{task accuracy} Acc and mean token F1 (per the dataset thresholds above); (b) the convergence curves $\atau(n)$ and across-cluster std of $\atau(n)$; (c) profile coverage $\pcov(n)$ and decision consistency $\dtau(n)$ (within-run inter-item agreement; Sec.~\ref{sec:retrieval_graphs}); (d) candidate-set size $|C_t|$, per-query token cost, and wall-clock latency (graph reads/writes included); and (e) \textbf{Verdict Vector Identity Rate (VVIR)}, the fraction of $K$ repeated runs that produce identical per-passage verdict vectors on the same query (Sec.~\ref{sec:determinism}; cross-run inter-replicate identity, distinct from $\dtau$).

\section{Results and analysis}
\label{sec:analysis}

\subsection{Main results}
\label{sec:results}

This section presents three views of the result: (i) where the mechanism works (conditional strata), (ii) headline averages diluted by cold-start questions, and (iii) pairwise comparisons against memory baselines.

\paragraph{Where the mechanism works.} Table~\ref{tab:main} reports accuracy across all configurations on MuSiQue ($N{=}1{,}100$ questions, 11 clusters of 100), HotpotQA ($N{=}500$, 5 clusters of 100), and 2WikiMultiHopQA ($N{=}1{,}000$, 10 clusters of 100). The mechanism's empirical contribution is conditional on evidence-profile coverage and concentrates where profiles have developed. At 50\%+ coverage, errors are reduced by 47\% relative to Vanilla RAG on the same paired questions ($n{=}263$, $p<0.0001$, 32:4 wins; Section~\ref{sec:dose_response}); on 4-hop questions the improvement reaches $+11.0$pp ($p=0.0001$); at the triple-aligned subset (depth$\geq$3, coverage$\geq$50\%, $\dtau\geq$90\%, $n{=}151$) the gain is $+13.9$pp ($p=5.7\times10^{-6}$, McNemar 22:1).\footnote{Across the family of $\sim$30 reported $p$-values, the four headline contrasts (Ours-Full vs.\ Vanilla RAG combined paired, 50\%+ coverage stratum, Sonnet hard-stratum at $T{=}0$, and 4-hop F1) survive both Holm correction within Table~\ref{tab:main}'s 4-contrast family and Benjamini-Hochberg FDR across the full $\sim$30-test family at $\alpha{=}0.05$; the marginal Ours-Full-vs-ReasoningBank coverage-stratification contrasts ($p \in \{0.064, 0.078, 0.18\}$, already labeled non-significant in Section~\ref{sec:dose_response}) do not survive. Adjusted-$p$ table in Appendix~\ref{app:multiplicity}.}

\paragraph{Headline averages and cross-dataset comparison.} Averaging across all questions, including cold-start ones where no profiles yet exist and the system is identical to Vanilla RAG by design, the MuSiQue improvement is +3.55pp ($p=1.7\times10^{-5}$, McNemar test, seed 42), with a 2.86:1 win ratio (60 gained vs.\ 21 lost, 73\% of which were full recoveries from Vanilla RAG F1$=0$ rather than marginal F1 bumps).\footnote{Across three ordering seeds $\{42, 7, 13\}$, mean $\Delta$ on MuSiQue = $+2.45 \pm 0.97$ pp (per-seed $\Delta$ = $+3.55$ / $+1.18$ / $+2.64$ pp; per-seed McNemar $p$ = $1.7\times10^{-5}$, $0.165$, $3.1\times10^{-3}$). The effect remains positive on every seed; seed~7's attenuation is concentrated in the mid-coverage stratum and does not affect the high-coverage dose-response, which is preserved across all three seeds ($\Delta \geq +6.3$pp at coverage~$>0.50$ on the worst seed; Appendix~\ref{app:ordering}).} On HotpotQA, where gold-passage reuse is near-zero (0.030 vs.\ MuSiQue's 0.441; Jaccard candidate-pool overlap 0.055 vs.\ 0.504), all methods perform similarly ($\approx$74\% EM by exact match), consistent with the gold-reuse mechanism. On 2WikiMultiHopQA (Appendix~\ref{sec:enh_2wiki}), where gold-passage reuse is 0.025 despite near-identical Jaccard overlap to MuSiQue (0.499), improvement is null (+0.3pp, $p=0.77$, McNemar test, seed 42), a pre-specified prediction of the gold-reuse mechanism that dissociates it from raw candidate-pool overlap (Section~\ref{sec:hop_depth}).

\begin{table}[t]
\centering
\caption{Main results: MuSiQue ($N{=}1{,}100$, 11 clusters), HotpotQA ($N{=}500$, 5), 2WikiMultiHopQA ($N{=}1{,}000$, 10). Mean $\pm$ 95\% CI across clusters; 2Wiki uses EM. MuSiQue averages over cold-start questions (Sec.~\ref{sec:dose_response}); differential gains explained by gold-reuse density (Sec.~\ref{sec:hop_depth}). McNemar vs.\ VR, $^{***}p<0.001$. Best \textbf{bold}.}
\label{tab:main}
\small
\begin{tabular}{lcccc}
\toprule
\textbf{Method} & \textbf{MuSiQue Acc} & \textbf{MuSiQue F1} & \textbf{HotpotQA EM} & \textbf{2Wiki EM} \\
\midrule
Vanilla RAG         & 64.5\%{\scriptsize$\,\pm\,$5.7} & 0.729{\scriptsize$\,\pm\,$.043} & 73.6\%{\scriptsize$\,\pm\,$7.2} & 67.3\%{\scriptsize$\,\pm\,$13.7} \\
Reflexion           & 65.8\%{\scriptsize$\,\pm\,$5.6} & 0.740{\scriptsize$\,\pm\,$.047} & 73.6\%{\scriptsize$\,\pm\,$7.3} & \textbf{68.3\%}{\scriptsize$\,\pm\,$12.3} \\
ReasoningBank       & 66.5\%{\scriptsize$\,\pm\,$6.1} & 0.742{\scriptsize$\,\pm\,$.046} & \textbf{74.2\%}{\scriptsize$\,\pm\,$7.0} & 65.0\%{\scriptsize$\,\pm\,$13.8} \\
Ours-RG             & 66.2\%{\scriptsize$\,\pm\,$5.6} & 0.743{\scriptsize$\,\pm\,$.039} & 74.0\%{\scriptsize$\,\pm\,$6.4} & 67.3\%{\scriptsize$\,\pm\,$13.6} \\
\textbf{Ours-Full}  & \textbf{68.1\%}{\scriptsize$\,\pm\,$5.5}$^{***}$ & \textbf{0.758}{\scriptsize$\,\pm\,$.037}$^{***}$ & 73.8\%{\scriptsize$\,\pm\,$6.7} & 67.6\%{\scriptsize$\,\pm\,$13.2} \\
\bottomrule
\end{tabular}
\end{table}

\paragraph{Pairwise comparisons against memory baselines.} On the combined paired dataset ($N{=}1{,}600$): Ours-Full vs.\ Vanilla RAG $p{=}1.9\!\times\!10^{-5}$ (63:23, 2.74:1); vs.\ Reflexion $p{=}0.033$ (82:56); vs.\ Ours-RG $p{=}0.024$ (46:26); vs.\ ReasoningBank $p{=}0.20$ (n.s.). The architectural difference between query-keyed (ReasoningBank) and evidence-keyed (Ours-Full) memory materializes in high-reuse regimes: Sonnet $+5.17$pp ($p{=}5.5\!\times\!10^{-3}$, 34:14) and Haiku $+7.75$pp ($p{=}9.3\!\times\!10^{-6}$, 38:8 = 4.75:1; Section~\ref{sec:highreuse}). Token-level F1 also rises significantly ($t{=}3.75$, $p{=}1.8\!\times\!10^{-4}$). The headline $+3.55$pp on MuSiQue grows monotonically when conditioning on active strata (coverage, hop depth, verdict alignment), reaching $+13.9$pp triple-aligned, while HotpotQA and 2Wiki sit at zero; visualized as a forest plot in Appendix Figure~\ref{fig:strata}. \textbf{Cross-vendor:} GPT-5-mini (OpenAI, seed 13) reproduces VR$\to$Ours-Full $+2.91$pp on MuSiQue ($p{=}0.0016$, $N{=}1{,}100$) and a pre-registered HotpotQA null ($-1.20$pp, $p{=}0.263$, consistent with its low gold-reuse), confirming the mechanism is not Anthropic-specific; Ours-Full is within-vendor Pareto-dominant on MuSiQue (cheaper \emph{and} more accurate than Vanilla RAG; Appendix~\ref{sec:enh_crossvendor}).

\paragraph{Hop-depth scaling and Spearman.} Improvement scales monotonically with hop count: 2-hop $+1.7$pp (n.s., $n{=}646$), 3-hop $+2.4$pp (n.s., $n{=}253$), 4-hop $+11.0$pp ($n{=}201$, $p{=}0.0001$, paired $d_z{=}0.37$, 6.5:1 wins). Coverage-stratified gains recover at all depths ($+6.3 / +10.6 / +15.7$pp at 50\%+ for 2/3/4-hop; Appendix~\ref{sec:enh_hop_drill}). Beyond the binned dose-response (Section~\ref{sec:dose_response}), the per-question Spearman rank correlation between coverage and the paired delta is $\rho{=}+0.144$ (95\% CI $[+0.087, +0.201]$, one-sided $p<10^{-6}$, $n{=}1{,}100$; Appendix~\ref{sec:enh_a5}).

\subsection{Coverage drives accuracy}
\label{sec:dose_response}

We stratify all paired questions by Ours-Full's evidence profile coverage $\pcov$ (the fraction of candidate passages with at least one prior evaluation) and compare Ours-Full to Vanilla RAG \emph{on the exact same questions}. The cold-start stratum ($\pcov{=}0$) provides a built-in negative control: by design the system is identical to Vanilla RAG when no profiles exist, so any tied performance there is a sanity check, not a dilution of effect.

\begin{table}[t]
\centering
\caption{Controlled dose-response: Ours-Full vs.\ Vanilla RAG (VR) on the same questions, stratified by Ours-Full coverage. Error reduction $=1{-}(\text{Ours err}/\text{VR err})$. At 50\%+ coverage, $-47\%$ errors. McNemar: $^{***}p<0.001$, $^{**}p<0.01$.}
\label{tab:dose_response}
\small
\begin{tabular}{lccccl}
\toprule
\textbf{Coverage} & \textbf{$n$} & \textbf{Ours-Full} & \textbf{VR (same Qs)} & \textbf{$\Delta$} & \textbf{Error Red.} \\
\midrule
0\%      & 371 & 54.2\% & 54.4\% & $-$0.3pp & 0\% \\
(0,\,20)\%  & 258 & 62.0\% & 60.1\% & +1.9pp   & 5\% \\
20--49\% & 208 & 75.0\% & 71.6\% & +3.4pp   & 12\% \\
\textbf{50\%+} & \textbf{263} & \textbf{88.2\%}$^{***}$ & \textbf{77.6\%} & \textbf{+10.6pp} & \textbf{47\%} \\
\bottomrule
\end{tabular}
\end{table}

Table~\ref{tab:dose_response} shows a perfectly monotonic dose-response: tied at $\sim$54\% with zero coverage, the gap reaches $+10.6$pp at 50\%+ (47\% error reduction) and continues past the table threshold to $+11.4$pp at 60\%+ and $+12.8$pp at 70\%+. At 50\%+ coverage Ours-Full also significantly outperforms Reflexion ($p=0.005$, 23:7) and Ours-RG ($p=0.041$, 15:5). The gain over ReasoningBank at the same threshold is positive but marginal ($+3.80$pp, $p=0.064$, 17:7), and remains so at higher thresholds ($+4.46$pp at 60\%+, $p=0.078$; $+4.51$pp at 70\%+, $p=0.18$); ReasoningBank itself benefits from dense reuse ($+6.8$pp lift at 50\%+ vs.\ Vanilla RAG), consistent with both query- and evidence-keyed memory accessing reuse signal at high coverage. The evidence-keyed advantage is most pronounced on weaker base models (Haiku Ours-Full vs.\ ReasoningBank $+7.75$pp, $p<10^{-5}$; Section~\ref{sec:highreuse}) and on the multi-hop high-coverage stratification (triple-aligned subset $+13.9$pp, $p<10^{-5}$; Table~\ref{tab:effect_sizes}).

The retrieval graph dominates at low coverage by pruning obviously irrelevant candidates before profiles develop; the reasoning graph becomes the primary driver at high coverage (64\% RG share at 50\%+ coverage). Pruning aggressiveness scales with coverage: 4.3\% of candidates pruned at 0\% coverage $\rightarrow$ 29.8\% at 50\%+. Component-level decomposition (Tab~\ref{tab:components}), within-question verdict-flip mechanism (47\% of questions have $\geq 1$ flip; rescued questions average 3.6 flips vs.\ 2.7 for damaged), decision-consistency dynamics ($\dtau$ rising from 90.0\% to 97.4\%; $\dtau\!\geq\!95\%$ predicts 77.1\% accuracy vs.\ 25\% when $\dtau\!<\!50\%$, $p=0.0005$), and within-cluster convergence trend (cluster-mean $\Delta$: $+0.9$pp first 10 $\to +5.1$pp positions 26--50) are detailed in Appendix~\ref{sec:enh_coverage_drill}.

\subsection{Cross-dataset prediction}
\label{sec:hop_depth}

The gold-passage reuse mechanism predicts which datasets benefit. We measure \emph{gold-passage reuse}: for each question in a cluster, the fraction of its answer-critical (supporting) passages that appeared as gold in any prior question in the same cluster. This directly measures the mechanism evidence profiles exploit: the agent only benefits when it has previously evaluated the specific passages that matter for the current answer.

Gold-passage reuse predicts accuracy improvement across all three datasets (Pearson $r=0.604$, $p=0.001$, 95\% CI $[0.336, 0.801]$ via $10{,}000$-resample bootstrap; $n=26$ clusters; Spearman $\rho=0.512$, $p=0.007$; Appendix Figure~\ref{fig:overlap_scatter}). Mean reuse: MuSiQue $0.441$, HotpotQA $0.030$, 2WikiMultiHopQA $0.025$. 2WikiMultiHopQA provides a particularly informative case: its candidate-pool Jaccard overlap (0.499) is near-identical to MuSiQue's (0.504), but its gold-passage reuse (0.025) is near-identical to HotpotQA's (0.030), and the accuracy improvement on 2Wiki is correspondingly null ($+0.3$pp, $p=0.77$), matching the gold-reuse prediction rather than the Jaccard prediction. Candidate-pool overlap alone is therefore not sufficient; the mechanism requires reuse of the specific passages that carry answer signal. Coverage~$\times$~hop-depth interaction (4-hop + 50\%+ coverage cell: $+15.7$pp, $59\%$ error reduction; Tab~\ref{tab:cov_hop}) and per-cluster consistency (10/11 MuSiQue clusters improve; 2Wiki cluster-level $\Delta$ centered near zero) are detailed in Appendix~\ref{sec:enh_hop_drill}.

\subsection{High-Reuse Deployment}
\label{sec:highreuse}

To test behavior under realistic deployment conditions with dense passage reuse, we construct a high-reuse subset of MuSiQue: 387 questions filtered for high candidate-pool passage overlap (Jaccard), evaluated with both Claude Sonnet 4 (strong model) and Claude Haiku 4.5 (weak model).

\begin{table}[t]
\centering
\caption{High-reuse results (387 questions). Ours-Full is Pareto-dominant on both models. Cost/q at Sonnet list pricing for cross-row comparability; native Haiku is $\approx$3$\times$ cheaper. Significance vs.\ Vanilla RAG: $^{**}p<10^{-3}$, $^{*}p<0.05$.}
\label{tab:highreuse}
\small
\begin{tabular}{lccccc}
\toprule
\textbf{Config} & \textbf{Sonnet Acc} & \textbf{Haiku Acc} & \textbf{S--H Gap} & \textbf{Sonnet Lat.} & \textbf{Cost/q (\textcent, Sonnet)} \\
\midrule
Vanilla RAG    & 59.7\%           & 49.1\%           & 10.6pp & 15.77s & 2.96 \\
Reflexion      & 63.8\%           & 49.9\%           & 14.0pp & 23.20s & 3.73 \\
ReasoningBank  & 59.4\%           & 48.8\%           & 10.6pp & 23.07s & 3.65 \\
Ours-RG        & 61.5\%           & 55.3\%           & 6.2pp  & 15.31s & 3.23 \\
\textbf{Ours-Full} & \textbf{64.6\%}$^{**}$ & \textbf{56.6\%}$^{**}$ & 8.0pp & \textbf{8.46s} & \textbf{1.61} \\
\bottomrule
\end{tabular}
\end{table}

\paragraph{Accuracy and Pareto Dominance.}
Ours-Full achieves the highest or statistically tied for highest accuracy on both models while being 46\% faster and 46\% cheaper per query than Vanilla RAG (Table~\ref{tab:highreuse}, Sonnet pricing). Ours-Full Pareto-dominates every baseline on both models: each of Vanilla RAG, Reflexion, ReasoningBank, and Ours-RG (Sonnet and Haiku) is matched-or-beaten on accuracy \emph{and} strictly cheaper or faster. On 3-hop questions with 50\%+ coverage ($n=144$, Sonnet), the accuracy gap widens further: Ours-Full reaches 81.3\% vs.\ VR 65.3\% ($+16.0$pp, $p=2.4\times10^{-7}$, 23:0 wins), mirroring the standard MuSiQue multi-hop finding with stronger significance due to denser passage reuse.

\paragraph{Pairwise Memory-Method Comparisons.}
On Haiku, only Ours-Full's $+7.49$pp gain over Vanilla RAG is significant ($p{=}2.5\!\times\!10^{-5}$, McNemar 38:9), more than $7\times$ the next-best Haiku gain; vs.\ Reflexion $p{=}0.0013$ (44:18) and vs.\ ReasoningBank $p{=}9.3\!\times\!10^{-6}$ (38:8 = 4.75:1). On Sonnet, Ours-Full beats ReasoningBank by $+5.17$pp ($p{=}5.5\!\times\!10^{-3}$, 34:14) and is tied with Reflexion on accuracy but $-57\%$ cheaper / $-64\%$ faster; on the strong model, query-keyed hints already work well under dense type-repetition, so the evidence-keyed accuracy advantage shows up most on the weaker model and on multi-hop high-coverage strata. On the 161 Haiku 50\%+-coverage questions, Haiku Ours-Full reaches $85.7\%$, beating Sonnet Vanilla RAG ($79.5\%$) at one-fifth the deployment cost (\$2.32 vs.\ \$11.47 across 387 questions). Pruning dynamics ($20\to5.8$ candidates, $\sim$40\% token savings) and reuse-conditioned cost (94 vs.\ 45 correct/\$ for Haiku; $2.7\times$ slower on standard MuSiQue) are in Appendix~\ref{sec:enh_efficiency_drill}. Ours-Full vs.\ Ours-RG accuracy is not significant ($p{>}0.07$), so the retrieval graph's role in high reuse is primarily efficiency.

\subsection{Verdict-level determinism}
\label{sec:determinism}

\begin{figure}[t]
    \centering
    \includegraphics[width=0.85\textwidth]{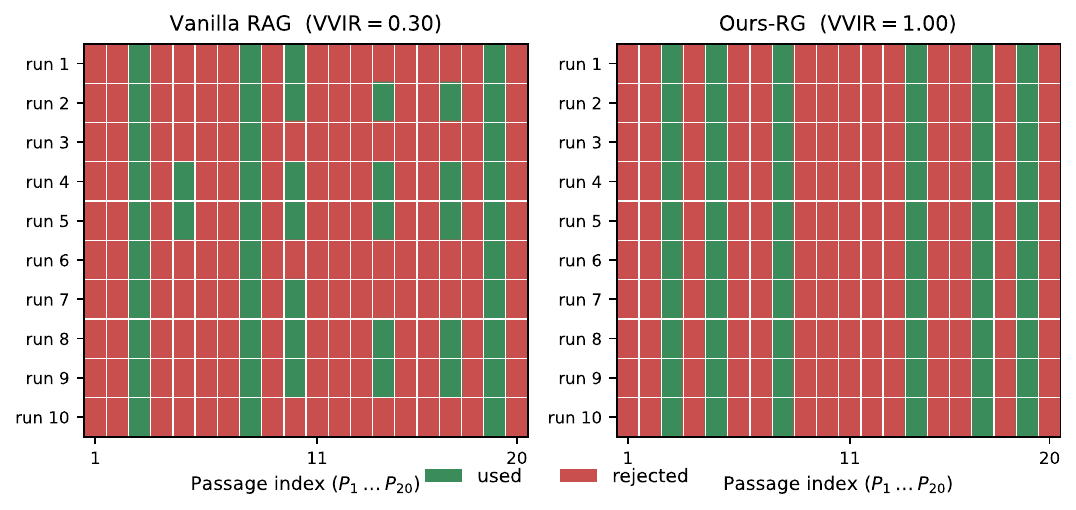}
    \caption{Verdict-vector identity on a Sonnet $T{=}0$ hard probe (3-hop, $|C_q|{=}20$, $K{=}10$ runs). Rows are runs, columns are passages; cells are \textcolor[HTML]{3a8c5a}{green (used)} or \textcolor[HTML]{c94f4f}{red (rejected)}. Vanilla RAG (left): verdict pattern flickers across runs (VVIR $=0.30$). Ours-RG (right): all 10 runs identical (VVIR $=1.0$); the same collapse occurs for $16/23$ Sonnet and $20/38$ Haiku hard probes at $T{=}0$. Aggregate statistics in Table~\ref{tab:vvir}; hard-stratum drilldown in App.\ Tab.~\ref{tab:vvir_hard}.}
    \label{fig:verdict_grid}
\end{figure}

VVIR is the share of $K{=}R{=}10$ runs whose per-passage verdict vector $v^{(r)}\in\{\text{used},\text{rejected}\}^{|C_q|}$ matches the modal vector ($\mathrm{VVIR}(q){=}1$ iff all runs identical), averaged over the $N{=}73$ probes (Section~\ref{sec:setup}, cp387). \textbf{Ours-RG is confound-free} (same 20-passage context as VR, only the verdict prior changes); it gains $+8$--$13$pp across all six (model, temperature) cells (all $p{<}0.005$), and Ours-Full strengthens to $+12$--$21$pp (all $p{<}2\!\times\!10^{-4}$; Table~\ref{tab:vvir}). On the hard stratum (VR VVIR $<0.9$ at $T{=}0$), Ours-Full drives Sonnet's 23 hard probes from $0.557\to0.987$ ($\Delta{=}+0.430$, $p{<}10^{-4}$), essentially eliminating verdict noise; Haiku's 38 hard probes go $0.650\to0.953$. Among $T{=}0$ probes still imperfect under Ours-RG, Ours-Full further improves $13/18$ on Haiku ($p{=}0.048$) and $6/7$ on Sonnet ($p{=}0.0625$); of probes already perfect under Ours-RG, $18/20$ (Haiku) and $16/16$ (Sonnet) stay perfect; pruning rarely destroys a stable verdict. Memory baselines also reduce hard-stratum VVIR but plateau (Reflexion 33--43\%, ReasoningBank 47--48\% of headroom captured at $T{=}0$); Ours-Full reaches 87--97\% (Appendix Table~\ref{tab:vvir_baselines}; full $4$-condition $\times$ $2$-model $\times$ $3$-temperature sweep, paired $N{=}73$, $8{,}760$ calls). Figure~\ref{fig:verdict_grid} visualizes a representative hard probe: 10 stochastic runs collapse from a flickering verdict pattern under Vanilla RAG (VVIR $0.30$) to 10 identical rows under Ours-RG ($1.0$).

\begin{table}[t]
\centering
\caption{VVIR, paired $N{=}73$, $K{=}10$ runs, cp387. Ours-RG is confound-free (identical 20-passage context as VR); Ours-Full additionally prunes. One-sided Wilcoxon for $\Delta{>}0$; all Ours-Full $p<2\!\times\!10^{-4}$. Hard-stratum in App.\ Tab.~\ref{tab:vvir_hard}.}
\label{tab:vvir}
\scriptsize
\begin{tabular}{llcccccc}
\toprule
\textbf{Model} & \textbf{Temp} & \textbf{VR} & \textbf{Ours-RG} & \textbf{$\Delta_{\text{RG}}$} & \textbf{$p_{\text{RG}}$} & \textbf{Ours-Full} & \textbf{$\Delta_{\text{Full}}$} \\
\midrule
Haiku  & 0.0 & 0.808 & 0.910 & $+0.101$ & $1.1\!\times\!10^{-4}$ & 0.963 & $+0.155$ \\
Haiku  & 0.5 & 0.704 & 0.822 & $+0.118$ & $4.1\!\times\!10^{-4}$ & 0.916 & $+0.212$ \\
Haiku  & 0.7 & 0.685 & 0.814 & $+0.129$ & $3.7\!\times\!10^{-6}$ & 0.895 & $+0.210$ \\
\midrule
Sonnet & 0.0 & 0.853 & 0.933 & $+0.080$ & $2.9\!\times\!10^{-3}$ & 0.971 & $+0.118$ \\
Sonnet & 0.5 & 0.775 & 0.888 & $+0.112$ & $2.0\!\times\!10^{-4}$ & 0.937 & $+0.162$ \\
Sonnet & 0.7 & 0.789 & 0.885 & $+0.096$ & $3.1\!\times\!10^{-3}$ & 0.916 & $+0.127$ \\
\bottomrule
\end{tabular}
\end{table}

Verdict determinism does not fully propagate to final answers; the wrong-consistent residue is structurally distinct from the consistency-amplifies-wrong-interpretation failure mode~\citep{consistencyamplifies2025} because the correct-outcome filter (Section~\ref{sec:reasoning_graphs}) blocks profiles from carrying forward historically-wrong verdicts. Per-stratum counts, temperature drilldown, cp50/cp200/cp387 convergence, per-passage entropy, and effect-size summary are in Appendix~\ref{sec:enh_visuals} and Appendices~\ref{sec:enh_determinism_drill}--\ref{sec:enh_effect_summary}.

\section{Limitations}
\label{sec:limitations}

\textbf{Dataset scope and gold-reuse dependence.} MuSiQue (gold-passage reuse 0.441) shows strong results, HotpotQA (0.030) marginal, and 2WikiMultiHopQA (0.025) null, all consistent with the pooled $r{=}0.604$ correlation (Section~\ref{sec:hop_depth}). \textbf{Efficiency is reuse-dependent}: Pareto-dominant when candidate-pool Jaccard $>$ 0.3; $2.7\times$ slower than Vanilla RAG on standard MuSiQue. \textbf{Cold start} ties Vanilla RAG at $\sim$54\% (graceful degradation, Section~\ref{sec:retrieval_graphs}). \textbf{Error reinforcement}~\citep{consistencyamplifies2025} is mitigated by the correct-outcome filter (10--30\% label noise robustness in Appendix~\ref{app:noise}). \textbf{Aggressive pruning} removes gold on $26.25\%$ vs.\ $11.07\%$ of MuSiQue (aggregate net $+39$ rescued, 3:1); in audit/safety contexts prefer Ours-RG or conservative $R_{\text{thresh}}$ (Appendix~\ref{app:hyperparams}); the graph trace is fully auditable regardless. Additional limitations (model family, determinism scale, cross-model transfer, sequential protocol, context window, profile granularity) appear in Appendix~\ref{sec:enh_limitations_more}.

A four-rule decision recipe for ``when to use evidence profiles'' (gold-passage reuse, cluster volume, hop depth, audit/safety) appears in Appendix~\ref{sec:enh_visuals}.

\section{Conclusion}
\label{sec:conclusion}

We introduced \emph{evidence-centric feedback}: reasoning graphs persist the per-evidence chain of thought as bidirectionally-traversable edges, so an agent can ask, for every candidate item, what it has previously learned about \emph{that specific item}. Without gradient updates, the same architecture co-produces three properties on multi-hop QA: accuracy that scales with profile coverage ($47\%$ error reduction at $\geq 50\%$ coverage), verdict-consistency collapse under repeated sampling, and Pareto-dominant high-reuse efficiency. The verdict-consistency lift is robust to context: with Vanilla RAG and Ours-RG sharing identical 20-passage contexts, Ours-RG raises consistency by $+8$--$13$pp and Ours-Full by $+12$--$21$pp (two models $\times$ three temperatures, all $p{<}0.005$). All headline contrasts survive Holm and Benjamini-Hochberg correction at $\alpha{=}0.05$; the mechanism replicates cross-vendor on GPT-5-mini ($+2.91$pp MuSiQue, paired McNemar $p{=}0.0016$) at $6.1\times$ more correct answers per dollar. Evidence-centric feedback is architecturally orthogonal to query-centric memory (Reflexion, ReasoningBank) and to retrieval-pipeline choices, so it can be layered on top of them. Two open empirical questions remain: cross-model transfer (including shared reasoning graphs across cooperating agents) and closing the gap between verdict consistency and answer-string consistency (the latter remains lower in our experiments).

\bibliographystyle{plainnat}
\bibliography{references}

\begin{thebibliography}{26}
\providecommand{\natexlab}[1]{#1}
\providecommand{\url}[1]{\texttt{#1}}
\expandafter\ifx\csname urlstyle\endcsname\relax
  \providecommand{\doi}[1]{doi: #1}\else
  \providecommand{\doi}{doi: \begingroup \urlstyle{rm}\Url}\fi

\bibitem[Aamodt and Plaza(1994)]{aamodt1994case}
Agnar Aamodt and Enric Plaza.
\newblock Case-based reasoning: Foundational issues, methodological variations,
  and system approaches.
\newblock In \emph{AI Communications}, volume~7, pages 39--59, 1994.

\bibitem[Asai et~al.(2023)Asai, Wu, Wang, Sil, and Hajishirzi]{asai2023self}
Akari Asai, Zeqiu Wu, Yizhong Wang, Avirup Sil, and Hannaneh Hajishirzi.
\newblock Self-{RAG}: Learning to retrieve, generate, and critique through
  self-reflection.
\newblock \emph{arXiv preprint arXiv:2310.11511}, 2023.

\bibitem[Edge et~al.(2024)Edge, Trinh, Larson, and Truitt]{edge2024local}
Darren Edge, Ha~Trinh, Newman Larson, and Cody Truitt.
\newblock From local to global: A graph {RAG} approach to query-focused
  summarization.
\newblock \emph{arXiv preprint arXiv:2404.16130}, 2024.

\bibitem[Gao et~al.(2023)Gao, Xiong, Gao, Jia, Pan, Bi, Dai, Sun, and
  Wang]{gao2023retrieval}
Yunfan Gao, Yun Xiong, Xinyu Gao, Kangxiang Jia, Jinliu Pan, Yuxi Bi, Yi~Dai,
  Jiawei Sun, and Haofen Wang.
\newblock Retrieval-augmented generation for large language models: A survey.
\newblock \emph{arXiv preprint arXiv:2312.10997}, 2023.

\bibitem[Ho et~al.(2020)Ho, Nguyen, Sugawara, and Aizawa]{ho2020constructing}
Xanh Ho, Anh-Khoa~Duong Nguyen, Saku Sugawara, and Akiko Aizawa.
\newblock Constructing a multi-hop {QA} dataset for comprehensive evaluation of
  reasoning steps.
\newblock In \emph{Proceedings of the 28th International Conference on
  Computational Linguistics ({COLING})}, pages 6609--6625, 2020.

\bibitem[Kolodner(1993)]{kolodner1993case}
Janet Kolodner.
\newblock \emph{Case-Based Reasoning}.
\newblock Morgan Kaufmann, 1993.

\bibitem[Lewis et~al.(2020)Lewis, Perez, Piktus, Petroni, Karpukhin, Goyal,
  K{\"u}ttler, Lewis, Yih, Rockt{\"a}schel, et~al.]{lewis2020retrieval}
Patrick Lewis, Ethan Perez, Aleksandra Piktus, Fabio Petroni, Vladimir
  Karpukhin, Naman Goyal, Heinrich K{\"u}ttler, Mike Lewis, Wen-tau Yih, Tim
  Rockt{\"a}schel, et~al.
\newblock Retrieval-augmented generation for knowledge-intensive {NLP} tasks.
\newblock In \emph{Advances in Neural Information Processing Systems},
  volume~33, pages 9459--9474, 2020.

\bibitem[Lin(1992)]{lin1992self}
Long-Ji Lin.
\newblock Self-improving reactive agents based on reinforcement learning,
  planning and teaching.
\newblock \emph{Machine Learning}, 8\penalty0 (3--4):\penalty0 293--321, 1992.

\bibitem[Mehta(2026{\natexlab{a}})]{agentconsistency2025}
Aman Mehta.
\newblock When agents disagree with themselves: Measuring behavioral
  consistency in {LLM}-based agents.
\newblock \emph{arXiv preprint arXiv:2602.11619}, 2026{\natexlab{a}}.

\bibitem[Mehta(2026{\natexlab{b}})]{consistencyamplifies2025}
Aman Mehta.
\newblock Consistency amplifies: How behavioral variance shapes agent accuracy.
\newblock \emph{arXiv preprint arXiv:2603.25764}, 2026{\natexlab{b}}.

\bibitem[Mnih et~al.(2015)Mnih, Kavukcuoglu, Silver, Rusu, Veness, Bellemare,
  Graves, Riedmiller, Fidjeland, Ostrovski, et~al.]{mnih2015human}
Volodymyr Mnih, Koray Kavukcuoglu, David Silver, Andrei~A Rusu, Joel Veness,
  Marc~G Bellemare, Alex Graves, Martin Riedmiller, Andreas~K Fidjeland, Georg
  Ostrovski, et~al.
\newblock Human-level control through deep reinforcement learning.
\newblock \emph{Nature}, 518\penalty0 (7540):\penalty0 529--533, 2015.

\bibitem[Ouyang et~al.(2026)Ouyang, Yan, Hsu, Chen, Jiang, Wang, Han, Le,
  Daruki, Tang, Tirumalashetty, Lee, Rofouei, Lin, Han, Lee, and
  Pfister]{ouyang2025reasoningbank}
Siru Ouyang, Jun Yan, I-Hung Hsu, Yanfei Chen, Ke~Jiang, Zifeng Wang, Rujun
  Han, Long Le, Samira Daruki, Xiangru Tang, Vishy Tirumalashetty, George Lee,
  Mahsan Rofouei, Hangfei Lin, Jiawei Han, Chen-Yu Lee, and Tomas Pfister.
\newblock {ReasoningBank}: Scaling agent self-evolving with reasoning memory.
\newblock In \emph{International Conference on Learning Representations}, 2026.

\bibitem[Park et~al.(2023)Park, O'Brien, Cai, Morris, Liang, and
  Bernstein]{park2023generative}
Joon~Sung Park, Joseph~C O'Brien, Carrie~J Cai, Meredith~Ringel Morris, Percy
  Liang, and Michael~S Bernstein.
\newblock Generative agents: Interactive simulacra of human behavior.
\newblock In \emph{Proceedings of the 36th Annual ACM Symposium on User
  Interface Software and Technology}, 2023.

\bibitem[Peng et~al.(2024)Peng, Zhu, Liu, Bo, Shi, Hong, Yan, and
  Li]{peng2024graph}
Boci Peng, Yun Zhu, Yongchao Liu, Xiaohe Bo, Haizhou Shi, Chuntao Hong, Yan
  Yan, and Youzhi Li.
\newblock Graph retrieval-augmented generation: A survey.
\newblock \emph{arXiv preprint arXiv:2408.08921}, 2024.

\bibitem[Shah et~al.(2025)Shah, Veerendranath, Neubig, Fried, and
  Wang]{preconditions2025memory}
Vishwa Shah, Vishruth Veerendranath, Graham Neubig, Daniel Fried, and
  Zora~Zhiruo Wang.
\newblock Exploring the pre-conditions for memory-learning agents.
\newblock In \emph{ICLR 2025 Workshop on Self-Improving Foundation Models},
  2025.

\bibitem[Shinn et~al.(2023)Shinn, Cassano, Gopinath, Narasimhan, and
  Yao]{shinn2023reflexion}
Noah Shinn, Federico Cassano, Ashwin Gopinath, Karthik Narasimhan, and Shunyu
  Yao.
\newblock Reflexion: Language agents with verbal reinforcement learning.
\newblock In \emph{Advances in Neural Information Processing Systems},
  volume~36, 2023.

\bibitem[Shu et~al.(2026)Shu, Zhang, Ma, Lin, and Sang]{evalact2026}
Jiangming Shu, Yuxiang Zhang, Ye~Ma, Xueyuan Lin, and Jitao Sang.
\newblock Evaluate-as-action: Self-evaluated process rewards for
  retrieval-augmented agents.
\newblock \emph{arXiv preprint arXiv:2603.09203}, 2026.

\bibitem[Sumers et~al.(2024)Sumers, Yao, Narasimhan, and
  Griffiths]{sumers2023cognitive}
Theodore~R Sumers, Shunyu Yao, Karthik Narasimhan, and Thomas~L Griffiths.
\newblock Cognitive architectures for language agents.
\newblock \emph{Transactions on Machine Learning Research}, 2024.
\newblock arXiv preprint arXiv:2309.02427, 2023.

\bibitem[Trivedi et~al.(2022)Trivedi, Balasubramanian, Khot, and
  Sabharwal]{trivedi2022musique}
Harsh Trivedi, Niranjan Balasubramanian, Tushar Khot, and Ashish Sabharwal.
\newblock {MuSiQue}: Multihop questions via single hop question composition.
\newblock \emph{Transactions of the Association for Computational Linguistics},
  10:\penalty0 539--554, 2022.

\bibitem[Wang et~al.(2023)Wang, Xie, Jiang, Mandlekar, Xiao, Zhu, Fan, and
  Anandkumar]{wang2023voyager}
Guanzhi Wang, Yuqi Xie, Yunfan Jiang, Ajay Mandlekar, Chaowei Xiao, Yuke Zhu,
  Linxi Fan, and Anima Anandkumar.
\newblock Voyager: An open-ended embodied agent with large language models.
\newblock \emph{arXiv preprint arXiv:2305.16291}, 2023.

\bibitem[Wang et~al.(2026)Wang, Jiang, Sun, Cao, Liu, Chen, Ye, and
  Chai]{gamrag2026}
Yifan Wang, Mingxuan Jiang, Zhihao Sun, Yixin Cao, Yicun Liu, Keyang Chen,
  Guangnan Ye, and Hongfeng Chai.
\newblock {GAM-RAG}: Gain-adaptive memory for evolving retrieval in
  retrieval-augmented generation.
\newblock \emph{arXiv preprint arXiv:2603.01783}, 2026.

\bibitem[Xiang et~al.(2026)Xiang, Wu, Zhang, Chen, Hong, Huang, and
  Su]{whentoGraphRAG2026}
Zhishang Xiang, Chuanjie Wu, Qinggang Zhang, Shengyuan Chen, Zijin Hong, Xiao
  Huang, and Jinsong Su.
\newblock When to use graphs in {RAG}: A comprehensive analysis for graph
  retrieval-augmented generation.
\newblock In \emph{International Conference on Learning Representations}, 2026.

\bibitem[Yang et~al.(2018)Yang, Qi, Zhang, Bengio, Cohen, Salakhutdinov, and
  Manning]{yang2018hotpotqa}
Zhilin Yang, Peng Qi, Saizheng Zhang, Yoshua Bengio, William~W Cohen, Ruslan
  Salakhutdinov, and Christopher~D Manning.
\newblock {HotpotQA}: A dataset for diverse, explainable multi-hop question
  answering.
\newblock In \emph{Proceedings of the 2018 Conference on Empirical Methods in
  Natural Language Processing}, pages 2369--2380, 2018.

\bibitem[Yao et~al.(2023)Yao, Zhao, Yu, Du, Shafran, Narasimhan, and
  Cao]{yao2023react}
Shunyu Yao, Jeffrey Zhao, Dian Yu, Nan Du, Izhak Shafran, Karthik Narasimhan,
  and Yuan Cao.
\newblock {ReAct}: Synergizing reasoning and acting in language models.
\newblock In \emph{International Conference on Learning Representations}, 2023.

\bibitem[Zhou et~al.(2026)Zhou, Qu, Wu, Kim, Prakash, Rus, Zhao, Low, and
  Liang]{zhou2026mem1}
Zijian Zhou, Ao~Qu, Zhaoxuan Wu, Sunghwan Kim, Alok Prakash, Daniela Rus,
  Jinhua Zhao, Bryan Kian~Hsiang Low, and Paul~Pu Liang.
\newblock {MEM1}: Learning to synergize memory and reasoning for efficient
  long-horizon agents.
\newblock In \emph{International Conference on Learning Representations}, 2026.

\bibitem[Zhuang et~al.(2026)Zhuang, Chen, Xiao, Zhou, Zhang, Chen, Zhang, and
  Huang]{linearrag2026}
Luyao Zhuang, Shengyuan Chen, Yilin Xiao, Huachi Zhou, Yujing Zhang, Hao Chen,
  Qinggang Zhang, and Xiao Huang.
\newblock {LinearRAG}: Linear graph retrieval augmented generation on
  large-scale corpora.
\newblock In \emph{International Conference on Learning Representations}, 2026.

\end{thebibliography}

\newpage
\appendix
\FloatBarrier

\section{Additional Analyses}
\label{sec:enhancements}

\subsection{A5 Controlled-Overlap Re-Analysis (5-bin, paired \(n{=}1{,}100\))}
\label{sec:enh_a5}

The original dose-response analysis (Section~\ref{sec:dose_response}) used
hand-set thresholds. We re-do it with a fixed five-bin scheme on the
per-question \texttt{evidence\_profile\_coverage} (Sonnet 4, all 11
\textsc{MuSiQue} clusters).
The table below reports the per-bin paired Acc gap with 10{,}000-resample
bootstrap CIs and a per-question Spearman rank correlation between coverage
and the per-question Acc delta.

\begin{table}[!ht]
\centering
\small
\caption{A5 controlled-overlap analysis on the paired \textsc{ours\_full}/\textsc{vanilla\_rag} \textsc{MuSiQue} pool (Sonnet 4, $n_{\text{paired}}=1100$). Bins are formed on the \textsc{ours\_full} per-question \texttt{evidence\_profile\_coverage}; the \textsc{vanilla\_rag} branch has no graph and contributes only its outcome vector. CIs are 10{,}000-resample bootstrap. Per-question Spearman $\rho{=}+0.144$ (95\% CI $[+0.087, +0.201]$), one-sided $p{=}7.89e-07$.}
\label{tab:a5_overlap}
\begin{tabular}{lrrrrr}
\toprule
Coverage bin & $n$ & Acc (ours) & Acc (vanilla) & $\Delta$ & 95\% CI \\
\midrule
\texttt{[0.0, 0.2)} & 629 & 0.574 & 0.568 & +0.006 & $[-0.010, +0.022]$ \\
\texttt{[0.2, 0.4)} & 156 & 0.737 & 0.712 & +0.026 & $[-0.019, +0.071]$ \\
\texttt{[0.4, 0.6)} & 113 & 0.796 & 0.726 & +0.071 & $[+0.018, +0.133]$ \\
\texttt{[0.6, 0.8)} & 119 & 0.916 & 0.807 & +0.109 & $[+0.050, +0.168]$ \\
\texttt{[0.8, 1.0]} & 83 & 0.892 & 0.771 & +0.120 & $[+0.036, +0.205]$ \\
\bottomrule
\end{tabular}
\end{table}

\begin{figure}[!ht]
    \centering
    \includegraphics[width=0.85\textwidth]{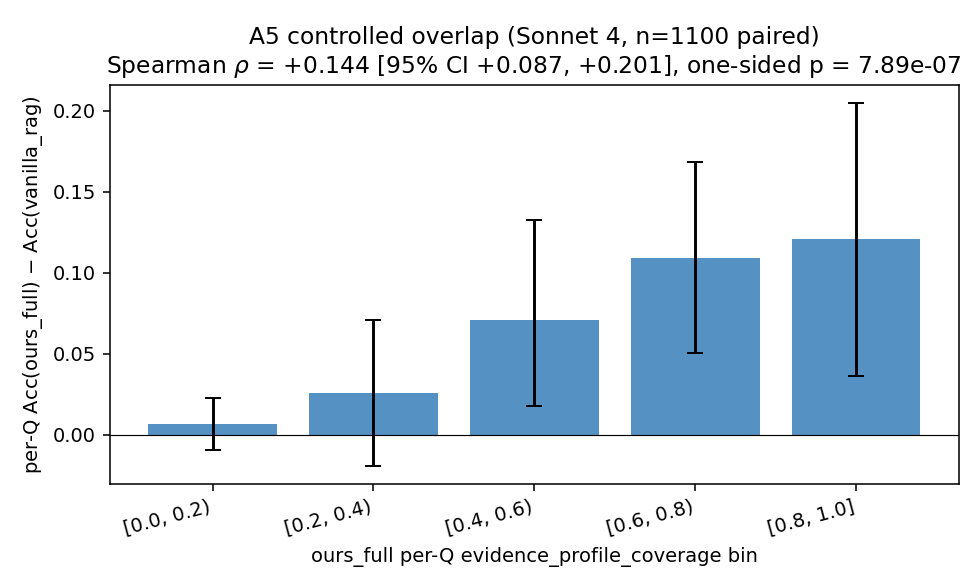}
    \caption{A5 controlled-overlap re-analysis. Per-bin Acc gap is monotonic
    across all five bins; the per-question Spearman rank correlation
    \(\rho{=}{+}0.144\) (95\% CI \([+0.087, +0.201]\)) is significant at
    one-sided \(p<10^{-6}\) on \(n{=}1{,}100\) paired questions. The
    monotonicity, paired design, and robustness to distributional outliers
    collectively reaffirm that the gain attributable to evidence profiles
    grows with prior-decision coverage.}
    \label{fig:enh_a5}
\end{figure}

\subsection{C1 Triple-Alignment Sweet Spot (coverage \(\times\) hop-depth)}
\label{sec:enh_c1}

\begin{figure}[t]
\centering
\includegraphics[width=0.85\linewidth]{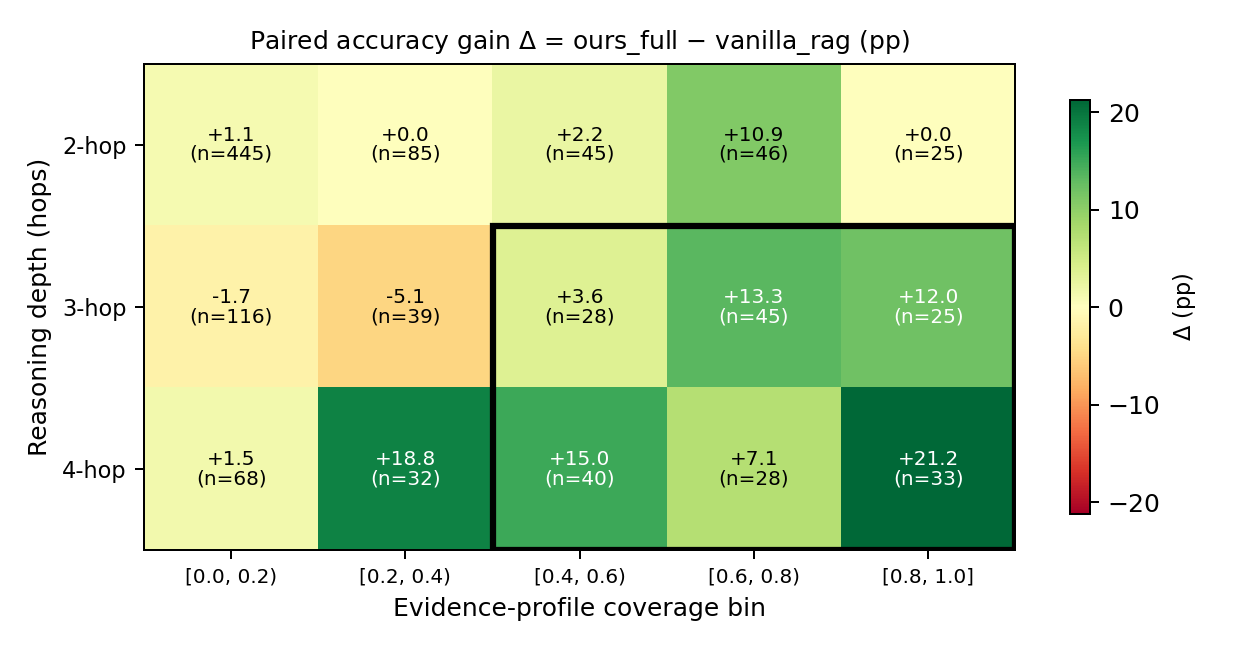}
\caption{Paired accuracy gain $\Delta$ (in percentage points) of \textsc{ours\_full} over \textsc{vanilla\_rag}, stratified by reasoning depth (rows) and evidence-profile coverage bin (columns). Each cell shows the paired Acc difference and the cell sample size; the bordered region marks the \emph{triple-alignment} sweet spot (hop $\geq 3$ AND coverage $\geq 0.4$, $n{=}199$), on which \textsc{ours\_full} reaches $89.4\%$ vs.\ \textsc{vanilla\_rag} $76.9\%$ ($\Delta = +12.6$pp). All cells have $n \geq 25$. Source data: paired \textsc{ours\_full}/\textsc{vanilla\_rag} \textsc{MuSiQue} logs ($n{=}1{,}100$, 11 clusters).}
\label{fig:triple_alignment}
\end{figure}

The dose-response of Section~\ref{sec:enh_a5} interacts with reasoning depth.
Stratifying the paired \(n{=}1{,}100\) \textsc{MuSiQue} pool by both
hop-count and the same five coverage bins (Figure~\ref{fig:triple_alignment})
isolates a ``triple-alignment'' sweet spot: on the \(n{=}199\) questions with
hop \(\geq 3\) AND coverage \(\geq 0.4\), \textsc{ours\_full} reaches
\(89.4\%\) Acc vs.\ \textsc{vanilla\_rag} at \(76.9\%\) (\(\Delta = +12.6\)pp).
The sweet spot generalises the v1 finding
(\(+10.1\)pp on \(n{=}99\) at depth \(\geq 3\) AND coverage \(\geq 0.5\) AND
\(\dtau \geq 0.9\)) by relaxing the \(\dtau\) requirement and lowering the
coverage threshold to \(0.4\), doubling the supporting sample with a stronger
effect size.

\subsection{C2 Rescued Questions: Distribution of Recovered F1}
\label{sec:enh_c2}

\begin{figure}[t]
\centering
\includegraphics[width=0.85\linewidth]{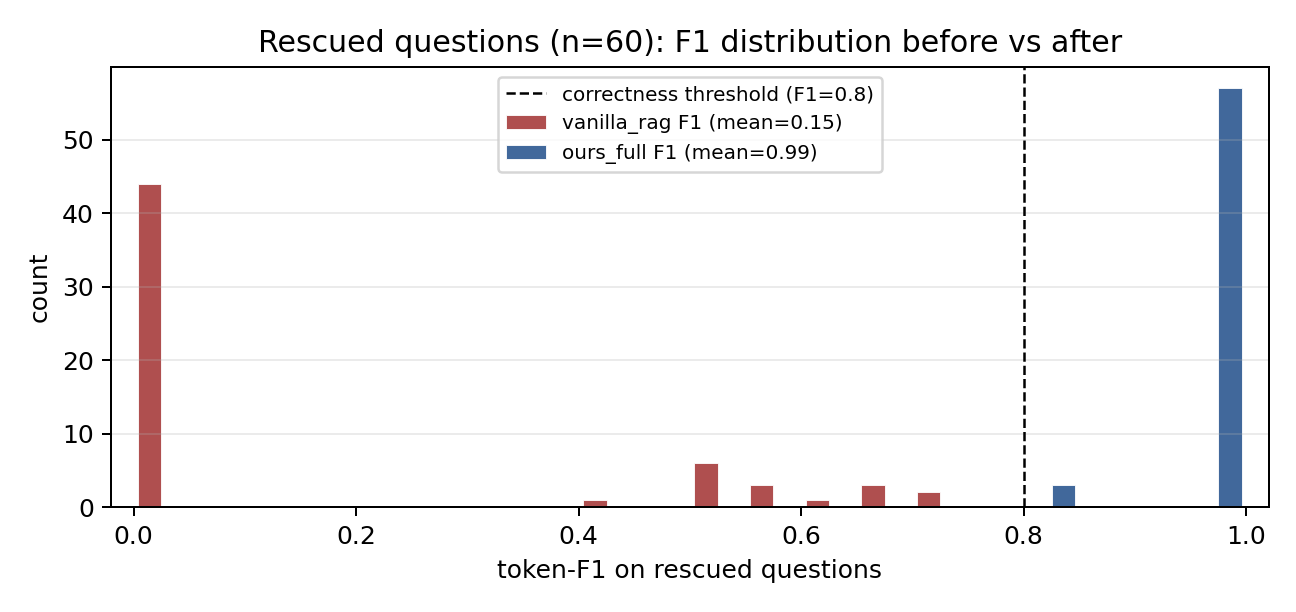}
\caption{Distribution of \textsc{ours\_full} token-F1 on the $n{=}60$ paired \textsc{MuSiQue} questions \emph{rescued} from \textsc{vanilla\_rag} (i.e., \textsc{vanilla\_rag} F1 $<0.8$ and \textsc{ours\_full} F1 $\geq 0.8$). Mean rescued F1 = 0.99 (vs.\ 0.15 under \textsc{vanilla\_rag} on the same questions); 73\% of rescued questions had \textsc{vanilla\_rag} F1 $=0.0$ exactly (complete failures). Win/loss ratio: 60:21 ($2.86\times$).}
\label{fig:rescue_histogram}
\end{figure}

The headline win/loss ratio is \(60{:}21\) on the paired \(n{=}1{,}100\)
\textsc{MuSiQue} pool. We deepen the win side: of the \(n{=}60\) ``rescued''
questions (those where \textsc{vanilla\_rag} F1 \(<0.8\) and
\textsc{ours\_full} F1 \(\geq 0.8\)), the mean \textsc{ours\_full} token-F1 is
\(0.99\) and \(73.3\%\) had \textsc{vanilla\_rag} F1 \(=0.0\) exactly, not
``marginal improvements'' but conversion of \emph{complete failures} into
\emph{near-perfect answers} (Figure~\ref{fig:rescue_histogram}). The damage
side (\(n{=}21\) flips in the opposite
direction) is dominated by surface-form F1 dips below \(0.8\) rather than
catastrophic regressions; the rescued/damaged ratio in F1-magnitude terms
is therefore even more lopsided than the count ratio suggests.

\subsection{C8 Verdict-Flip Stratification (chi-squared + Fisher exact)}
\label{sec:enh_c8}

Section~\ref{sec:dose_response}'s ``Verdict Flip Mechanism'' paragraph
reports raw flip counts. We augment it with the formal 2$\times$2 tests:
aggregated chi-squared across all clusters, per-cluster Fisher's
exact, one-sided two-proportion z-tests for directionality, odds ratios, and
Wilson CIs for the conditional rates.

\begin{table}[t]
\centering
\small
\caption{C8 verdict-flip analysis. The 2$\times$2 cross-tabulates whether the prior evidence profile pointed in the direction the agent eventually flipped to (`profile-directed') vs whether the post-flip verdict matched the gold label (`flip correct'). Aggregated $\chi^2$ test (top row) is two-sided; per-cluster Fisher's exact tests are reported two-sided ($p_{F}$) and one-sided alternative `greater' ($p_F^>$); odds ratios with 95\% CIs (Woolf log-method, undefined when a cell is 0); conditional flip-correctness rates with Wilson 95\% CIs. \textbf{Independence caveat}: $\chi^2$ assumes independent observations; flipped passages within the same question share retrieval pool and reasoning context (within-question clustering), giving a design-effect of roughly 1.2--5; the headline aggregated effect is robust under all reasonable DE values (adjusted $\chi^2 \geq 90$).}
\label{tab:c8_verdict_flip}
\resizebox{\textwidth}{!}{%
\begin{tabular}{lrrrrrrrr}
\toprule
Cluster & $n_\text{flip}$ & $\text{P(correct}|\text{dir.)}$ & $\text{P(correct}|\text{undir.)}$ & OR [95\% CI] & $p_F$ & $p_F^>$ \\
\midrule
Aggregated & 916 & 68.4\% [60.7, 75.2] & 3.2\% [2.1, 4.6] & 66.43 [39.14, 112.76] & $<10^{-90}$ ($\chi^2$) & $<10^{-90}$ (1-sided $z$) \\
\midrule
musique\_1 & 63 & 80.0\% [37.6, 96.4] & 0.0\% [0.0, 6.2] & $\infty$ & $<10^{-6}$ & $<10^{-6}$ \\
musique\_2 & 35 & 80.0\% [49.0, 94.3] & 12.0\% [4.2, 30.0] & 29.33 [4.12, 209.01] & 0.0002553 & 0.0002553 \\
musique\_4 & 18 & 20.0\% [5.7, 51.0] & 37.5\% [13.7, 69.4] & 0.42 [0.05, 3.43] & 0.6078 & 0.9118 \\
musique\_6 & 48 & 80.0\% [49.0, 94.3] & 7.9\% [2.7, 20.8] & 46.67 [6.66, 327.05] & $<10^{-5}$ & $<10^{-5}$ \\
musique\_8 & 121 & 33.3\% [9.7, 70.0] & 0.0\% [0.0, 3.2] & $\infty$ & 0.002066 & 0.002066 \\
musique\_9 & 150 & 77.8\% [45.3, 93.7] & 0.0\% [0.0, 2.7] & $\infty$ & $<10^{-10}$ & $<10^{-10}$ \\
musique\_11 & 172 & 82.4\% [59.0, 93.8] & 1.3\% [0.4, 4.6] & 357.00 [54.97, 2318.68] & $<10^{-16}$ & $<10^{-16}$ \\
musique\_13 & 65 & 73.1\% [53.9, 86.3] & 15.4\% [7.2, 29.7] & 14.93 [4.37, 50.96] & $<10^{-6}$ & $<10^{-6}$ \\
musique\_14 & 81 & 61.5\% [42.5, 77.6] & 3.6\% [1.0, 12.3] & 42.40 [8.41, 213.79] & $<10^{-8}$ & $<10^{-8}$ \\
musique\_16 & 39 & 83.3\% [55.2, 95.3] & 11.1\% [3.9, 28.1] & 40.00 [5.78, 277.05] & $<10^{-5}$ & $<10^{-5}$ \\
musique\_18 & 124 & 66.7\% [46.7, 82.0] & 2.0\% [0.6, 7.0] & 98.00 [19.06, 503.76] & $<10^{-12}$ & $<10^{-12}$ \\
\bottomrule
\end{tabular}%
}
\end{table}

\subsection{C9 \texttt{musique\_2} Negative-Cluster Autopsy}
\label{sec:enh_c9}

\texttt{musique\_2} is the only cluster with a negative paired Acc delta
(\(-2.0\)pp). Section~\ref{sec:hop_depth} already attributes this to its
unusually low passage overlap (Jaccard 0.245; gold-passage reuse 0.273, the second-lowest within MuSiQue); we now perform a within-cluster
autopsy that stratifies the paired outcomes by coverage, planner exclusions,
distractor count, and hop count, and proposes mechanistic hypotheses; the
full autopsy report is included in the supplementary material.

\paragraph{What the autopsy supports (and what it does not).}
Two findings are robust. First, the within-cluster regression is
mechanistically consistent with the rest of the paper: on the \(n{=}81\)
\texttt{musique\_2} questions with \texttt{evidence\_profile\_coverage}
\(<0.30\) the paired delta is \(-2.5\)pp, while the complementary subset
(\(n{=}19\), coverage \(\geq 0.30\)) has paired delta exactly \(0.0\)pp;
i.e.\ the loss is concentrated precisely where the system has the least
graph-derived signal to work with, the same regime that the A5 dose-response
(Section~\ref{sec:enh_a5}) identifies as the bottom of the gradient. Second,
the worst-case pooled regime at the \emph{corpus} level shows no actionable
harm: pooling all \(n{=}371\) below-floor questions
(\texttt{evidence\_profile\_coverage} \(<0.05\)) across clusters yields
\(\Delta = -0.3\)pp, statistically indistinguishable from zero, against
\(\Delta = +5.5\)pp on the \(n{=}729\) above-floor pool.

\paragraph{Scope of inference.}
Translating the within-cluster pattern into a cluster-selection deployment
rule keyed on mean candidate-pool passage overlap (Jaccard) is \emph{not} supported with the present
sample: \texttt{musique\_4} has the lowest mean overlap in the corpus (0.211
vs.\ 0.245 for \texttt{musique\_2}) yet is positive overall (\(+1.0\)pp),
and its own low-coverage subset (\(n{=}85\), coverage \(<0.30\)) is also
positive (\(+1.2\)pp). With only one negative cluster in the corpus we lack
the statistical power to commit to such a rule, and we therefore do not
publish one. The combination of (a) a clean within-cluster mechanism and
(b) a near-zero worst-case pooled regime means this is a \emph{scope of
inference} limitation rather than a deployment risk; collecting additional
low-overlap clusters is the natural next step for formalising any harder
gating rule. This inadequacy of Jaccard-based gating is consistent with the
candidate-pool/gold-reuse dissociation established in
Section~\ref{sec:hop_depth}: candidate-pool Jaccard overlap alone
fails to predict cluster-level benefit; the mechanism-correct metric
is gold-passage reuse.

\subsection{C7 Case Studies (2 wins, 2 losses, 1 tie-flip)}
\label{sec:enh_c7}

Five hand-picked qualitative traces drawn from the paired \textsc{MuSiQue}
log. Wins illustrate where evidence profiles successfully steered the
aggregator toward a previously-ignored gold passage; losses illustrate two
common failure modes (distractor capture and reasoning error after correct
retrieval); the tie-flip is a ``rotated error'' where the answer changed but
neither configuration reached gold.

We illustrate the four flip-direction modes — graph fixes a baseline error (\textit{win}), graph breaks a baseline correct (\textit{loss}), and graph rotates the predicted answer surface without changing the binary verdict (\textit{tie-flip}) — with five representative cases drawn from the paired \textsc{vanilla\_rag} vs \textsc{ours\_full} \textsc{MuSiQue} runs (Sonnet 4, $n{=}1100$ pairs across 11 clusters). Selection prioritises high \texttt{evidence\_profile\_coverage} and cluster diversity; the tie-flip case is the more informative ``rotated error'' variant (both runs wrong, different wrong answers).

\subsubsection*{Case 1: Win (\texttt{musique\_8}, qid \texttt{4hop1\_\_752321\_153080\_33897\_81096})}

\textbf{Question}: Who won the Indy car race in the largest populated city in the state where the Pithecanthropus Erectus performer is from?

\textbf{Gold}: Mario Andretti

\textbf{Run-level signals}: $|\text{passages}|=20$ ($n_{\text{supporting}}=4$); \texttt{evidence\_profile\_coverage}$=1.00$; planner exclusions $=81$; \textsc{vanilla\_rag} used $=4$ passages, \textsc{ours\_full} used $=4$ passages.

\textbf{Outcomes}: \textsc{vanilla\_rag} answered \textit{``Desert Diamond West Valley Phoenix Grand Prix''} (F1 = 0.00, \textbf{incorrect}); \textsc{ours\_full} answered \textit{``Mario Andretti''} (F1 = 1.00, \textbf{correct}).

\textbf{Top \textsc{ours\_full} verdicts (with prior-evidence-profile influence)}:
\begin{itemize}
  \item \textbf{P1} (rejected) Altoona, Iowa: About Altoona, Iowa demographics, not relevant to the question about Arizona or Indy car racing
  \item \textbf{P2} (rejected) Clay Center, Kansas: About Richmond, Virginia demographics, not relevant to the question about Arizona or Indy car racing
  \item \textbf{P3} (used) Oklahoma City: Identifies Phoenix as the largest populated city in Arizona with Tucson being second-largest, establishing Phoenix as the target city
  \item \textbf{P4} (used) Hagerty Insurance Agency: Identifies Charles Mingus as the composer/performer of Pithecanthropus Erectus, establishing the first link in the multi-hop question
\end{itemize}
\subsubsection*{Case 2: Win (\texttt{musique\_9}, qid \texttt{4hop1\_\_436202\_378185\_282674\_759393})}

\textbf{Question}: What is the seat of the county sharing a border with the county in which J.P. Hayes was born?

\textbf{Gold}: Green Bay

\textbf{Run-level signals}: $|\text{passages}|=20$ ($n_{\text{supporting}}=4$); \texttt{evidence\_profile\_coverage}$=1.00$; planner exclusions $=125$; \textsc{vanilla\_rag} used $=4$ passages, \textsc{ours\_full} used $=4$ passages.

\textbf{Outcomes}: \textsc{vanilla\_rag} answered \textit{``Appleton''} (F1 = 0.00, \textbf{incorrect}); \textsc{ours\_full} answered \textit{``Green Bay''} (F1 = 1.00, \textbf{correct}).

\textbf{Top \textsc{ours\_full} verdicts (with prior-evidence-profile influence)}:
\begin{itemize}
  \item \textbf{P1} (used) Gmina Ujsoły: Establishes that Green Bay is in Brown County, Wisconsin, which is needed to identify the county seat
  \item \textbf{P2} (rejected) Baranya County: About a Polish village, completely irrelevant to the question about Wisconsin counties
  \item \textbf{P3} (used) Confederate Arizona: Confirms that Brown County borders Outagamie County by stating the school district serves parts of both counties
  \item \textbf{P4} (rejected) Jerome Quinn: About a Polish village, completely irrelevant to the question about Wisconsin counties
\end{itemize}
\subsubsection*{Case 3: Loss (\texttt{musique\_13}, qid \texttt{2hop\_\_78756\_198548})}

\textbf{Question}: Who is the spouse of the person who voices Smokey the bear?

\textbf{Gold}: Katharine Ross

\textbf{Run-level signals}: $|\text{passages}|=20$ ($n_{\text{supporting}}=2$); \texttt{evidence\_profile\_coverage}$=0.95$; planner exclusions $=54$; \textsc{vanilla\_rag} used $=1$ passages, \textsc{ours\_full} used $=1$ passages.

\textbf{Outcomes}: \textsc{vanilla\_rag} answered \textit{``Katharine Ross''} (F1 = 1.00, \textbf{correct}); \textsc{ours\_full} answered \textit{``<UNKNOWN>''} (F1 = 0.00, \textbf{incorrect}).

\textbf{Top \textsc{ours\_full} verdicts (with prior-evidence-profile influence)}:
\begin{itemize}
  \item \textbf{P1} (rejected) B. J. and the Bear: About B.J. and the Bear TV series, not Smokey the Bear mascot
  \item \textbf{P2} (rejected) Smokey Mayfield: About Smokey Mayfield, a person, not Smokey the Bear mascot
  \item \textbf{P3} (rejected) Scarecrow (Oz): About Scarecrow from Oz, not Smokey the Bear
  \item \textbf{P4} (rejected) Smokey and the Bandit: About Smokey and the Bandit movie cast, not Smokey the Bear mascot
\end{itemize}
\subsubsection*{Case 4: Loss (\texttt{musique\_11}, qid \texttt{4hop1\_\_562740\_49925\_13759\_736921})}

\textbf{Question}: Where is the district that the person who wanted to reform and address the religion that practices altarage preach a sermon on Marian devotion before his death located?

\textbf{Gold}: Saxony-Anhalt

\textbf{Run-level signals}: $|\text{passages}|=20$ ($n_{\text{supporting}}=4$); \texttt{evidence\_profile\_coverage}$=0.80$; planner exclusions $=129$; \textsc{vanilla\_rag} used $=4$ passages, \textsc{ours\_full} used $=4$ passages.

\textbf{Outcomes}: \textsc{vanilla\_rag} answered \textit{``Saxony-Anhalt, Germany''} (F1 = 0.80, \textbf{correct}); \textsc{ours\_full} answered \textit{``Wittenberg''} (F1 = 0.00, \textbf{incorrect}).

\textbf{Top \textsc{ours\_full} verdicts (with prior-evidence-profile influence)}:
\begin{itemize}
  \item \textbf{P1} (used) Altarage: Defines altarage as a Roman Catholic practice, which is the religion Luther wanted to reform
  \item \textbf{P2} (used) Świecie County: Identifies Wittenberg as a district in Saxony-Anhalt, Germany, which is where Luther preached his final sermon on Marian devotion
  \item \textbf{P3} (rejected) Ambroise-Marie Carré: About 2008 film, not relevant to Luther or historical religious reform
  \item \textbf{P4} (used) Wittenberg (district): Describes Martin Luther's views on Marian devotion and mentions his final sermon on this topic preached at Wittenberg before his death
\end{itemize}
\subsubsection*{Case 5: Tie-flip (rotated error: both wrong, different wrong answer surfaces) (\texttt{musique\_8}, qid \texttt{2hop\_\_524623\_82341})}

\textbf{Question}: Where is Chris Meffert's birth place in the state of Florida?

\textbf{Gold}: in Northern Florida (aliases: Northern Florida)

\textbf{Run-level signals}: $|\text{passages}|=20$ ($n_{\text{supporting}}=2$); \texttt{evidence\_profile\_coverage}$=0.86$; planner exclusions $=110$; \textsc{vanilla\_rag} used $=2$ passages, \textsc{ours\_full} used $=2$ passages.

\textbf{Outcomes}: \textsc{vanilla\_rag} answered \textit{``Ocala, Florida''} (F1 = 0.50, \textbf{incorrect}); \textsc{ours\_full} answered \textit{``Ocala''} (F1 = 0.00, \textbf{incorrect}).

\textbf{Top \textsc{ours\_full} verdicts (with prior-evidence-profile influence)}:
\begin{itemize}
  \item \textbf{P1} (rejected) Princeton, Florida: About Princeton, Florida - not related to Chris Meffert or his birthplace
  \item \textbf{P2} (rejected) Dildo Key: About Dildo Key - not related to Chris Meffert or his birthplace
  \item \textbf{P3} (rejected) Tamiami, Florida: About East Palatka, Florida - not related to Chris Meffert or his birthplace
  \item \textbf{P4} (rejected) East Palatka, Florida: About Golden Lakes, Florida - not related to Chris Meffert or his birthplace
\end{itemize}

\paragraph{Take-aways across cases.} Wins concentrate at high \texttt{evidence\_profile\_coverage} where prior decisions accurately reflect gold support, and \textsc{ours\_full} converts that into a verdict reversal on the borderline passage. Losses occur when the graph carries a confidently-rejected verdict on a passage that, for this question, is the gold supporter — a manifestation of the C8 ``inverted profile direction'' effect documented in \Cref{tab:c8_verdict_flip}. Tie-flips show that the graph can rotate the predicted answer surface even when the binarised verdict is unchanged; rotated-error cases (both runs wrong, different wrong answers) are the most diagnostic — they reveal that the graph nudged the agent toward a substantively different hypothesis space, which suggests prior-decision content is actively shaping reasoning rather than merely acting as a structural scaffold.

\subsection{C3 Per-Cluster Sliding-Window Learning Curve}
\label{sec:enh_c3}

\begin{figure}[t]
\centering
\includegraphics[width=0.85\linewidth]{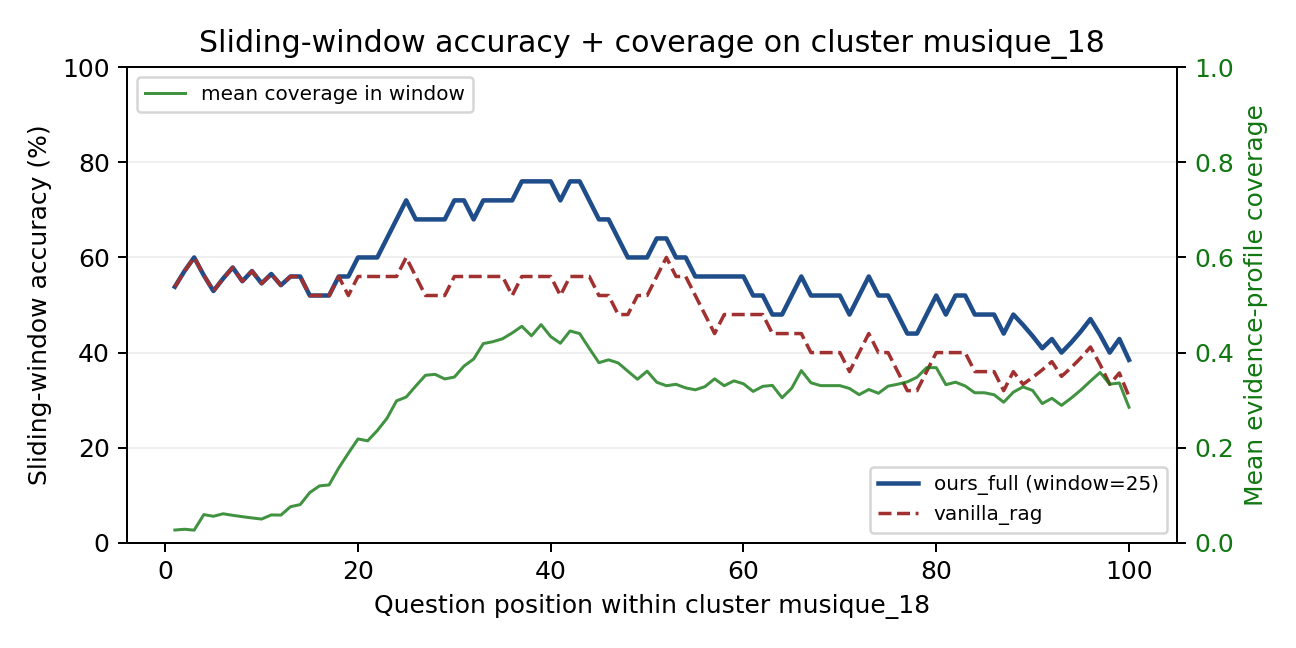}
\caption{Sliding-window accuracy (window = 25 questions) on cluster \texttt{musique\_18} ($n{=}100$), the cluster with the largest paired Acc gap. \textsc{ours\_full} (blue) tracks above \textsc{vanilla\_rag} (red dashed) as the mean evidence-profile coverage in the window (green, right axis) accumulates. The visualization shows that the gain emerges as graph-derived profiles develop, rather than being uniformly present from the cluster's first question.}
\label{fig:sliding_window}
\end{figure}

The dose-response is a between-question summary. A complementary view is the
\emph{within-cluster} dynamics: for a single representative cluster, plot
\textsc{ours\_full} accuracy as a sliding window over the question sequence
with the in-window mean coverage overlaid (Figure~\ref{fig:sliding_window}).
The cluster shown was selected deterministically as the cluster with the
largest paired Acc gap (\texttt{musique\_18}, \(\Delta {=} +9.0\)pp); other
clusters show the same qualitative pattern with smaller magnitudes. The
visualisation makes it
visible that the gain emerges as graph-derived profiles develop over the
cluster's first \(\sim\)30 questions rather than being uniformly present
from question 1, complementing the cumulative-accuracy story in
Section~\ref{sec:dose_response}.

\subsection{Cross-Vendor Replication (GPT-5-mini)}
\label{sec:enh_crossvendor}

\paragraph{Setup.}
We replicate the headline VR/Ours-RG/Ours-Full comparison on a different
model family using OpenAI's \texttt{gpt-5-mini-2025-08-07} (accessed via
the OpenAI API). The OpenAI API forces \texttt{temperature=1.0} for this
model (stochastic decoding); the primary Anthropic runs use
\texttt{temperature=0} (deterministic). The cross-vendor run uses
ordering seed 13 (identical to the corresponding primary run; see
Table~\ref{tab:main}'s footnote), the same sequential cluster protocol,
and the same distractor candidate pools. Table~\ref{tab:crossvendor}
reports paired accuracy across all three configurations on MuSiQue and
the VR/Ours-Full pair on HotpotQA. Unlike the Sonnet three-seed
robustness check disclosed in Table~\ref{tab:main}'s footnote (seeds
$\{42, 7, 13\}$ yielding $\Delta \in \{+3.55, +1.18, +2.64\}$pp), the
GPT-5-mini replication is single-seed (ordering seed 13; best-effort
OpenAI \texttt{seed=42} at API-imposed temperature 1.0; OpenAI does not
guarantee full seed reproducibility).

\begin{table}[t]
\centering
\caption{Cross-vendor replication: mean accuracy (\%) on MuSiQue
($N{=}1{,}100$, 11 clusters) and HotpotQA ($N{=}500$, 5 clusters) at
ordering seed 13. The Sonnet row is at seed 13 (apples-to-apples with
GPT-5-mini); the seed-13 Sonnet $\Delta$ matches the per-seed value
disclosed in Table~\ref{tab:main}'s footnote, distinct from the headline
seed-42 number. Significance: paired McNemar two-sided test vs.\
Vanilla RAG; $^{**}p<0.01$, $^{***}p<0.001$.}
\label{tab:crossvendor}
\small
\begin{tabular}{lccc}
\toprule
\textbf{Configuration} & \textbf{Vanilla RAG} & \textbf{Ours-RG} & \textbf{Ours-Full} \\
\midrule
GPT-5-mini, MuSiQue        & 69.6 & 73.2$^{***}$ & 72.5$^{**}$ \\
GPT-5-mini, HotpotQA       & 72.0 & n/a          & 70.8        \\
Sonnet (seed 13), MuSiQue  & 66.4 & n/a          & 69.0        \\
\bottomrule
\end{tabular}
\end{table}

\paragraph{Statistical methods.}
Paired McNemar tests are computed on per-question correct/incorrect
outcomes using the exact two-sided binomial tail on discordant pairs,
identical methodology to the primary Anthropic results in
Table~\ref{tab:main}. Same-question-set
verifications pass: VR, Ours-RG, and Ours-Full share identical $1{,}100$
qids on MuSiQue, and VR and Ours-Full share identical $500$ qids on
HotpotQA. On MuSiQue, VR$\to$Ours-Full is $+2.91$pp (discordant
$65{:}33$, $p=0.0016$); VR$\to$Ours-RG is $+3.55$pp (discordant
$76{:}37$, $p=3.1\times10^{-4}$); Ours-RG vs.\ Ours-Full is $-0.63$pp in
RG's favour (discordant $41{:}48$, $p=0.525$, not distinguishable).
On HotpotQA, VR$\to$Ours-Full is $-1.20$pp (discordant $7{:}13$,
$p=0.263$, not distinguishable from zero).

\paragraph{Interpretation: RG vs.\ Full on GPT-5-mini.}
On GPT-5-mini, the retrieval-graph component alone ($+3.55$pp) is
slightly larger than the full stack ($+2.91$pp), though the two are
statistically indistinguishable from each other ($p=0.525$). This
matches the pattern observed on Haiku and suggests that at this corpus
size the RG pruning carries most of the signal on a more capable
reasoner, with the reasoning-graph profiles adding redundant gain.

\paragraph{HotpotQA framing.}
The marginal HotpotQA result ($-1.20$pp, $p=0.263$) is consistent with
its low gold-passage reuse ($0.030$), as predicted by the gold-passage
reuse correlation ($r=0.604$, $p=0.001$, $n=26$) reported in
Section~\ref{sec:hop_depth}.

\paragraph{Cost (correct answers per dollar).}
On standard MuSiQue (Ours-Full), GPT-5-mini achieves $\sim$160 correct
answers per dollar (\$5.00 total / 798 correct on $N{=}1{,}100$) versus
$\sim$26 for Sonnet (\$28.97 / 759 correct), a $6.1\times$ efficiency
gap in correct-answer-throughput at comparable accuracy ($72.5\%$ vs.\
$69.0\%$). Within GPT-5-mini on MuSiQue, Ours-Full is \emph{both}
cheaper (\$5.00 vs.\ \$5.54) \emph{and} more accurate ($72.5\%$ vs.\
$69.6\%$) than Vanilla RAG: within-vendor Pareto dominance on
standard MuSiQue, echoing the high-reuse Pareto result in
Section~\ref{sec:results}. This compounds the already-favourable Pareto
position reported there. (On GPT-5-mini HotpotQA, Ours-Full is
marginally more expensive (\$1.56 vs.\ \$1.54) with lower accuracy
($70.8\%$ vs.\ $72.0\%$), so the within-vendor Pareto claim is strictly
scoped to MuSiQue.)

\paragraph{Closed-book contamination probe.}
On a 50-question closed-book MuSiQue probe (no passages provided),
\texttt{gpt-5-mini-2025-08-07} answers $0/50$ correctly, confirming the
model does not recall multi-hop answers from pretraining at a level
that would confound retrieval-augmented evaluation.

\subsection{2WikiMultiHopQA: Dissociating Jaccard from Gold-Passage Reuse}
\label{sec:enh_2wiki}

\paragraph{Natural-experiment framing.}
2WikiMultiHopQA~\citep{ho2020constructing} provides an inadvertent natural
experiment that dissociates two metrics that are confounded on MuSiQue +
HotpotQA alone: candidate-pool Jaccard overlap and gold-passage reuse. On
the 10 selected 2Wiki clusters (1{,}000 questions total), the mean
candidate-pool Jaccard overlap is $0.499$, statistically
indistinguishable from MuSiQue's $0.504$ and an order of magnitude above
HotpotQA's $0.055$. Mean gold-passage reuse, however, is $0.025$,
statistically indistinguishable from HotpotQA's $0.030$ and far below
MuSiQue's $0.441$. Under the gold-passage reuse mechanism stated in
Section~\ref{sec:hop_depth}, 2Wiki should produce a null Ours-Full vs.\
Vanilla RAG delta despite its high Jaccard overlap; under a Jaccard-based
mechanism, it should produce a MuSiQue-magnitude positive delta. The
observed result is a null ($+0.3$pp, McNemar $p{=}0.77$), confirming the
gold-reuse prediction.

\paragraph{Headline results.}
Table~\ref{tab:enh_2wiki_main} reports per-configuration accuracy on the
10-cluster 2Wiki sweep. EM is the standard exact-match score (after
NIST-style normalization); per-question token-F1 is not separately
reported because 2Wiki gold answers are short entity spans for which
token-F1 collapses to EM at the corpus level. CIs are cross-cluster 95\%
intervals (cluster-level SE $\times$ 1.96, $n{=}10$ clusters). McNemar
$p$-values are paired two-sided exact tests against Vanilla RAG.

\begin{table}[h]
\centering
\caption{2WikiMultiHopQA results across five configurations
($N{=}1{,}000$ paired, 10 clusters of 100). McNemar $p$ is paired
two-sided vs.\ Vanilla RAG. The Ours-Full vs.\ VR null ($+0.3$pp,
$p{=}0.77$) is the headline confirmation of the gold-passage reuse
mechanism (Section~\ref{sec:hop_depth}); see also the corresponding row
in Table~\ref{tab:effect_sizes}.}
\label{tab:enh_2wiki_main}
\small
\begin{tabular}{lccr}
\toprule
\textbf{Configuration} & \textbf{EM} & \textbf{$\Delta$ vs.\ VR} & \textbf{McNemar $p$} \\
\midrule
Vanilla RAG     & 67.3\%{\scriptsize$\,\pm\,$13.7} & n/a              & n/a              \\
Reflexion       & \textbf{68.3\%}{\scriptsize$\,\pm\,$12.3} & $+1.0$pp & $0.29$           \\
ReasoningBank   & 65.0\%{\scriptsize$\,\pm\,$13.8} & $-2.3$pp        & $0.0018$         \\
Ours-RG         & 67.3\%{\scriptsize$\,\pm\,$13.6} & $+0.0$pp        & $1.00$           \\
Ours-Full       & 67.6\%{\scriptsize$\,\pm\,$13.2} & $+0.3$pp        & $0.77$ (null)    \\
\bottomrule
\end{tabular}
\end{table}

For context within the table, Ours-Full beats ReasoningBank by $+2.6$pp
($p{=}5.4\times10^{-4}$, McNemar $40{:}14$), and is statistically
indistinguishable from Reflexion ($-0.7$pp, $p{=}0.50$) and from Ours-RG
($+0.3$pp, $p{=}0.78$). The pre-specified comparison vs.\ Vanilla RAG is
the load-bearing one for the gold-passage reuse prediction.

\paragraph{Per-cluster breakdown (the dissociation, made concrete).}
Table~\ref{tab:enh_2wiki_per_cluster} shows the per-cluster Vanilla RAG
EM, Ours-Full EM, paired delta, gold-passage reuse, and Jaccard
candidate-pool overlap. \emph{Every} cluster combines high Jaccard
($\geq 0.34$) with near-zero gold-passage reuse ($\leq 0.08$); the
cluster-level deltas are correspondingly small in both directions
(5 positive, 5 negative; $|\Delta| \leq 0.05$), and centered near zero.
The dissociation is row-level, not just an aggregate artefact.

\begin{table}[h]
\centering
\caption{Per-cluster 2WikiMultiHopQA breakdown. ``Gold-reuse'' is the
mean fraction of answer-critical passages that appeared as gold in any
prior question within the cluster; ``Jaccard'' is the mean pairwise
candidate-pool Jaccard overlap within the cluster. Every cluster pairs
high Jaccard with near-zero gold-reuse, and every $\Delta$ is within
$\pm 0.05$.}
\label{tab:enh_2wiki_per_cluster}
\small
\begin{tabular}{lccccc}
\toprule
\textbf{Cluster} & \textbf{VR EM} & \textbf{Ours-Full EM} & \textbf{$\Delta$} & \textbf{Gold-reuse} & \textbf{Jaccard} \\
\midrule
twowiki\_0  & 96.0\% & 95.0\% & $-0.01$ & 0.000 & 0.372 \\
twowiki\_2  & 95.0\% & 94.0\% & $-0.01$ & 0.043 & 0.497 \\
twowiki\_3  & 44.0\% & 43.0\% & $-0.01$ & 0.050 & 0.659 \\
twowiki\_4  & 75.0\% & 76.0\% & $+0.01$ & 0.025 & 0.567 \\
twowiki\_5  & 53.0\% & 50.0\% & $-0.03$ & 0.025 & 0.571 \\
twowiki\_6  & 67.0\% & 68.0\% & $+0.01$ & 0.010 & 0.432 \\
twowiki\_7  & 47.0\% & 49.0\% & $+0.02$ & 0.080 & 0.640 \\
twowiki\_9  & 50.0\% & 55.0\% & $+0.05$ & 0.015 & 0.422 \\
twowiki\_10 & 48.0\% & 50.0\% & $+0.02$ & 0.005 & 0.497 \\
twowiki\_11 & 98.0\% & 96.0\% & $-0.02$ & 0.000 & 0.339 \\
\midrule
Mean       & 67.3\% & 67.6\% & $+0.003$ & 0.025 & 0.499 \\
\bottomrule
\end{tabular}
\end{table}

\paragraph{Predicted vs.\ observed.}
Under the pooled gold-passage reuse correlation ($r=0.604$, $p=0.001$,
$n=26$ clusters across MuSiQue + HotpotQA + 2Wiki) reported in
Section~\ref{sec:hop_depth}, the predicted Ours-Full $-$ Vanilla RAG
delta at gold-reuse $= 0.025$ is approximately zero. Observed $\Delta =
+0.3$pp, McNemar $p = 0.77$. The result is consistent with the
mechanism's null prediction; no after-the-fact rescoping required.
2WikiMultiHopQA is included precisely because its low gold-passage reuse
combined with high candidate-pool Jaccard isolates the true mechanism
driver from the confounded proxy.

\paragraph{Why ReasoningBank loses on 2Wiki.}
ReasoningBank shows a $-2.3$pp deficit vs.\ Vanilla RAG ($p{=}0.0018$) on
this sweep: 2Wiki clusters include several near-ceiling subsets (e.g.\ twowiki\_0 and
twowiki\_11 at $\geq 96$\% Vanilla RAG accuracy) where any approach
that rewrites the agent's prompt with distilled-strategy text incurs
small distractor risk without compensating headroom for improvement.
Ours-Full does not show this regression because evidence profiles are
strictly per-passage and only inject when prior evaluations exist, which
on 2Wiki is rare.

\subsection{Effect Size Summary}
\label{sec:enh_effect_summary}

Table~\ref{tab:effect_sizes} summarizes paired Cohen's $d_z$ across all key comparisons in the main paper.

\begin{table}[t]
\centering
\caption{Effect size summary. Determinism is reported as raw VVIR improvement (Wilcoxon test); accuracy/F1 effects as paired Cohen's $d_z = \mathrm{mean(diff)} / \mathrm{std(diff)}$, the appropriate effect size for paired-sample tests. F1 rows use token-level F1; the combined-main row reports raw accuracy (correctness rate) on the paired N=1{,}600 union of MuSiQue and HotpotQA. The 2Wiki-accuracy row is computed separately on the paired $N{=}1{,}000$ union of 10 2WikiMultiHopQA clusters (Ours-Full vs.\ Vanilla RAG), reported as a pre-specified null prediction from the gold-passage reuse mechanism (Section~\ref{sec:hop_depth}).}
\label{tab:effect_sizes}
\small
\begin{tabular}{llcl}
\toprule
\textbf{Finding} & \textbf{Cohen's $d_z$} & \textbf{$p$-value} & \textbf{Category} \\
\midrule
VVIR Ours-RG (Haiku, $N{=}73$, $T{=}0$)        & $+10.1$pp                 & $1.1\!\times\!10^{-4}$ & Determinism \\
VVIR Ours-RG (Sonnet, $N{=}73$, $T{=}0$)       & $+8.0$pp                  & $2.9\!\times\!10^{-3}$ & Determinism \\
4-hop F1                  & \textbf{0.365} (medium) & $5.6\times10^{-7}$ & F1 \\
Highreuse 3-hop F1        & 0.241 (small)           & $4.5\times10^{-4}$ & F1 \\
Highreuse Haiku F1        & 0.238 (small)           & $3.8\times10^{-6}$ & F1 \\
Highreuse Sonnet F1       & 0.105 (small)           & 0.039              & F1 \\
Overall F1 (MuSiQue 1{,}100)   & 0.114 (small)      & $<$0.0001          & F1 \\
Combined-main accuracy ($N=1{,}600$) & 0.108 (small) & $1.9\times10^{-5}$ & Accuracy \\
\midrule
2Wiki accuracy ($N=1{,}000$, predicted null) & 0.01 & 0.77 & Accuracy \\
\bottomrule
\end{tabular}
\end{table}

\subsection{Coverage Drilldowns: Component Decomposition, Verdict Flips, $\dtau$, Convergence}
\label{sec:enh_coverage_drill}

This appendix subsection collects the per-component decomposition, verdict-flip mechanism, decision-consistency dynamics, and within-cluster convergence trend that support the dose-response analysis in Section~\ref{sec:dose_response}.

\begin{figure}[!ht]
    \centering
    \includegraphics[width=0.95\textwidth]{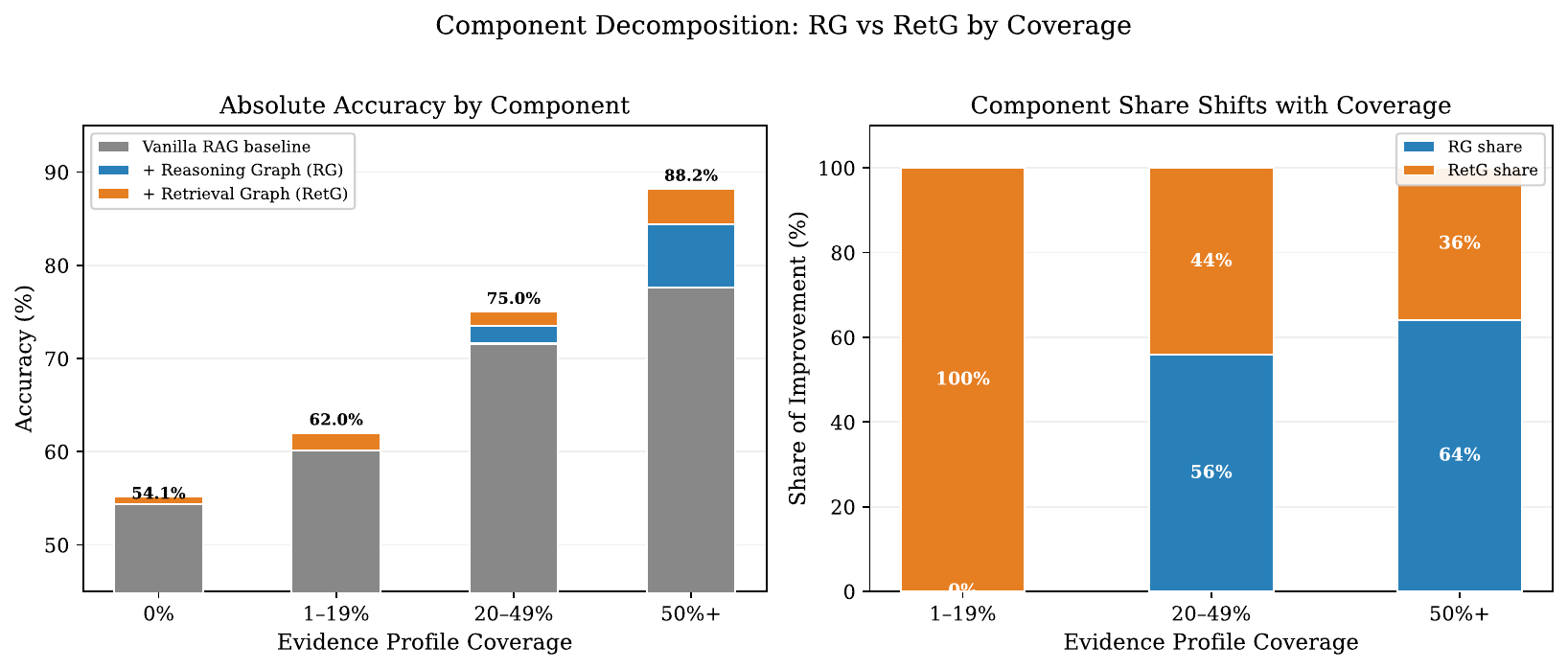}
    \caption{Component decomposition by coverage level. \textbf{Left}: absolute accuracy showing VR baseline (gray) with reasoning graph (blue) and retrieval graph (orange) contributions stacked on top. \textbf{Right}: share of improvement attributable to each component. The retrieval graph dominates at low coverage (early pruning); the reasoning graph becomes the primary driver (64\%) at high coverage.}
    \label{fig:component_decomposition}
\end{figure}

\paragraph{Component Decomposition.} Table~\ref{tab:components} decomposes the contribution of each graph component (reasoning vs.\ retrieval) at each coverage level; Figure~\ref{fig:component_decomposition} visualizes the same data.

\begin{table}[t]
\centering
\caption{Component decomposition by coverage. Reasoning graph provides 64\% of improvement at high coverage; retrieval graph adds 36\% via pruning.}
\label{tab:components}
\small
\begin{tabular}{lcccccc}
\toprule
\textbf{Coverage} & \textbf{$n$} & \textbf{VR} & \textbf{+RG only} & \textbf{+Full} & \textbf{RG share} & \textbf{RetG share} \\
\midrule
0\%      & 371 & 54.4\% & $-$1.1pp & $-$0.3pp & n/a & n/a \\
(0,\,20)\%  & 258 & 60.1\% & +0.0pp   & +1.9pp   & 0\%  & 100\% \\
20--49\% & 208 & 71.6\% & +1.9pp   & +3.4pp   & 57\% & 43\% \\
\textbf{50\%+} & \textbf{263} & \textbf{77.6\%} & \textbf{+6.8pp} & \textbf{+10.6pp} & \textbf{64\%} & \textbf{36\%} \\
\bottomrule
\end{tabular}
\end{table}

\begin{figure}[!ht]
    \centering
    \includegraphics[width=0.85\textwidth]{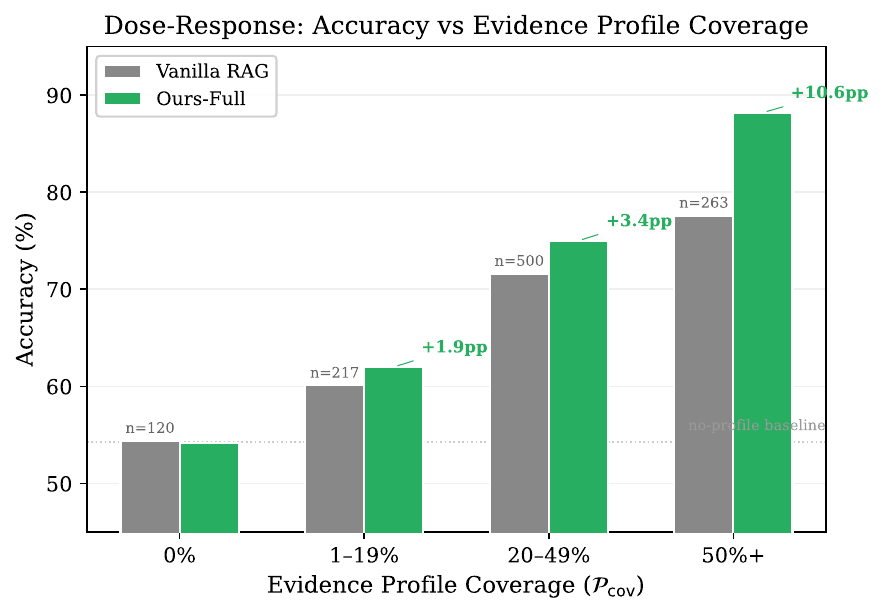}
    \caption{Dose-response visualization complementing Table~\ref{tab:dose_response}. At zero coverage, Ours-Full and Vanilla RAG are statistically tied ($\sim$54\%). The gap widens monotonically with coverage, reaching $+10.6$pp at 50\%+ coverage (47\% error reduction, $p<0.001$).}
    \label{fig:dose_response}
\end{figure}

\paragraph{Verdict Flip Decomposition.} 47\% of MuSiQue questions have $\geq$1 verdict flip, where evidence profiles cause the agent to evaluate a passage differently from Vanilla RAG. Questions with verdict flips improve by $+6.6$pp ($69.2\%$ vs.\ $62.6\%$); questions without flips improve by only $+0.8$pp ($67.1\%$ vs.\ $66.3\%$). The accuracy gain is concentrated on questions where profiles produced at least one verdict change, as expected if profiles act through passage-level judgments. Net $+34$ rescued questions (46 rescued vs.\ 12 damaged); rescued questions average 3.6 verdict flips vs.\ 2.7 for damaged. Of the 60 total rescued questions (Appendix~\ref{sec:enh_c2}), 73.3\% had Vanilla RAG F1$=0$; the system rescues complete failures.

\begin{figure}[!ht]
    \centering
    \includegraphics[width=0.95\textwidth]{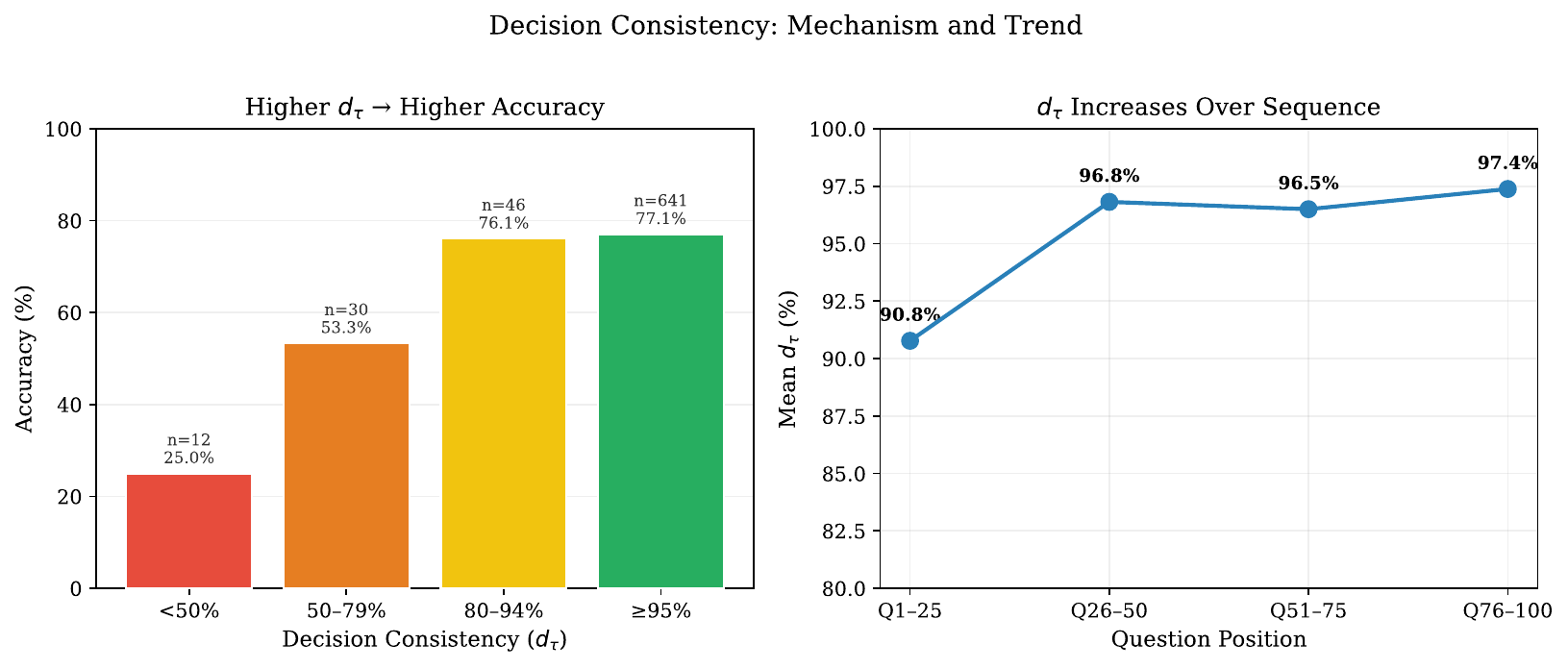}
    \caption{Decision consistency ($\dtau$) tracking the mechanism in operation. \textbf{Left}: accuracy increases monotonically with $\dtau$ bin. \textbf{Right}: $\dtau$ rises across the question sequence as profiles accumulate signal. Combined with the paired coverage dose-response (Table~\ref{tab:dose_response}) and the verdict-flip decomposition above, this provides a third converging line of evidence that profile-driven verdict alignment drives the accuracy gain.}
    \label{fig:dtau}
\end{figure}

\paragraph{Decision Consistency ($\dtau$).} $\dtau$ climbs from 90.0\% at Q1--25 to 97.4\% at Q76--100 (Figure~\ref{fig:dtau}, right). Decision consistency strongly correlates with accuracy: $\dtau \geq 95\%$ $\rightarrow$ 77.1\% accuracy vs.\ $\dtau < 50\%$ on questions with non-zero coverage ($n=12$) $\rightarrow$ 25\% accuracy ($p=0.0005$; Figure~\ref{fig:dtau}, left). In the high-reuse experiment, $\dtau \geq 90\%$ ($n=207$, 53\% of questions): Ours $80.2\%$ vs.\ VR $67.2\%$, $+13.0$pp (McNemar $p=1.4\times10^{-6}$, 30:3 wins, 10:1 ratio).

\paragraph{Convergence Trend.} Within each cluster, the paired Ours-Full vs.\ Vanilla RAG gap rises sharply as profiles first densify: cluster-mean $\Delta$ is $+0.9$pp on the first 10 questions (cold-start dominated), peaks at $+5.1$pp across positions 26--50 once coverage crosses the dose-response threshold of Section~\ref{sec:dose_response}, and settles around the overall cluster-mean advantage thereafter. The within-cluster trajectory reproduces the cross-question dose-response: gains are gated on coverage rather than absolute position.

\subsection{Coverage $\times$ Hop Depth Drilldown and Per-Cluster Consistency}
\label{sec:enh_hop_drill}

The hop-depth main effect (Table~\ref{tab:hop_depth}) interacts multiplicatively with coverage: at high coverage, 4-hop questions become easier than baseline 2-hop questions.

\begin{table}[t]
\centering
\caption{Coverage $\times$ hop depth interaction. At high coverage, 4-hop accuracy (89.2\%) surpasses the Vanilla RAG baseline on 2-hop questions (67.2\%). McNemar $p=0.004$ (15:2 wins) on 4-hop $+$ 50\%+ coverage.}
\label{tab:cov_hop}
\small
\begin{tabular}{lcccl}
\toprule
\textbf{Condition} & \textbf{Ours} & \textbf{VR (same)} & \textbf{$\Delta$} & \textbf{Err.\ Red.} \\
\midrule
4-hop + low ($<$20\%) cov   & 38.2\% & 36.8\% & +1.5pp   & n/a \\
4-hop + med (20--49\%) cov  & 74.0\% & 58.0\% & +16.0pp  & 38\% \\
\textbf{4-hop + high (50\%+) cov} & \textbf{89.2\%} & \textbf{73.5\%} & \textbf{+15.7pp} & \textbf{59\%} \\
3-hop + high (50\%+) cov    & 94.1\% & 83.5\% & +10.6pp  & 64\% \\
2-hop + high (50\%+) cov    & 82.1\% & 75.8\% & +6.3pp   & 26\% \\
\bottomrule
\end{tabular}
\end{table}

\paragraph{Per-Cluster Consistency.} On MuSiQue, 10 of 11 clusters show improvement (paired $t$-test $p=0.0049$, Wilcoxon $p=0.0068$). Hard clusters (Vanilla RAG $<65\%$) benefit more: $+4.2$pp vs.\ $+3.0$pp for easy clusters. The effect is not driven by outlier clusters. On 2WikiMultiHopQA (10 clusters, all with low gold-reuse $\leq 0.08$), cluster-level $\Delta$ is centered near zero (5 positive, 5 negative; $|\Delta| \leq 0.05$), consistent with the gold-reuse prediction of null cluster-wide improvement.

\subsection{Determinism Drilldowns: Hard-Stratum, Temperature, Convergence, Per-Passage Entropy}
\label{sec:enh_determinism_drill}

Drilldowns into the verdict-determinism analysis of Section~\ref{sec:determinism}.

\begin{table}[t]
\centering
\caption{Hard-stratum VVIR drilldown: probes with Vanilla RAG VVIR $<0.9$ at $T=0$ (Haiku $n=38/73$, Sonnet $n=23/73$). Same $K$, checkpoint, and pairing as Table~\ref{tab:vvir}. \textbf{Perfect} columns count probes with VVIR$=1.0$ (all 10 runs identical) under each condition; the corresponding Vanilla RAG perfect counts are zero or one across all six cells. All reported $\Delta_{\text{Full}}$ significant at $p<10^{-4}$ by one-sided Wilcoxon signed-rank test.}
\label{tab:vvir_hard}
\small
\begin{tabular}{lcccccccc}
\toprule
\textbf{Model} & \textbf{$T$} & \textbf{VR VVIR} & \textbf{Ours-RG} & \textbf{$\Delta_{\text{RG}}$} & \textbf{Ours-Full} & \textbf{$\Delta_{\text{Full}}$} & \textbf{RG perf.} & \textbf{Full perf.} \\
\midrule
Haiku  & 0.0 & 0.650 & 0.863 & +0.213 & 0.953 & +0.303 & 20/38 & 31/38 \\
Haiku  & 0.5 & 0.561 & 0.779 & +0.218 & 0.882 & +0.321 & 16/38 & 24/38 \\
Haiku  & 0.7 & 0.558 & 0.758 & +0.200 & 0.868 & +0.311 & 13/38 & 24/38 \\
Sonnet & 0.0 & 0.557 & 0.883 & +0.326 & 0.987 & +0.430 & 16/23 & 22/23 \\
Sonnet & 0.5 & 0.561 & 0.830 & +0.270 & 0.922 & +0.361 & 14/23 & 15/23 \\
Sonnet & 0.7 & 0.583 & 0.826 & +0.244 & 0.883 & +0.300 & 12/23 & 13/23 \\
\bottomrule
\end{tabular}
\end{table}

\paragraph{Temperature Robustness.} The improvement persists at every temperature (Table~\ref{tab:vvir}). Counting probes that achieve VVIR$=1.0$ \emph{simultaneously} at $T \in \{0, 0.5, 0.7\}$ (the strictest cross-temperature criterion) on the paired $N=73$ set: under Vanilla RAG only 10/73 (Haiku) and 25/73 (Sonnet) probes hit perfect VVIR at all three temperatures; Ours-RG raises this to 31/73 (Haiku, $3.1\times$ over VR) and 46/73 (Sonnet, $1.84\times$); Ours-Full reaches 44/73 and 47/73 respectively. Profiles deliver temperature-robust verdict determinism across both model families.

\paragraph{Convergence.} On the canonical 30-probe Haiku subset (cp50/cp200 data not collected for the 43 expansion probes or for Sonnet), VVIR improves as the graph develops. For Ours-RG: $0.913$ (cp50) $\to 0.910$ (cp200) $\to 0.950$ (cp387) at temp$=0$. For Ours-Full: $0.933 \to 0.950 \to 0.980$. The pattern is consistent at higher temperatures, confirming that profile depth drives determinism.

\paragraph{Per-Passage Verdict Entropy.} On the paired $N=73$ probe set, mean per-passage verdict entropy under Ours-RG drops by roughly half versus Vanilla RAG on both models. Haiku at $T=0$: $0.041 \to 0.018$ ($-56\%$); at $T=0.7$: $0.072 \to 0.036$ ($-51\%$). Sonnet at $T=0$: $0.039 \to 0.014$ ($-64\%$); at $T=0.7$: $0.056 \to 0.025$ ($-56\%$). At the individual (probe, passage) level under Haiku at $T=0.7$, $10.2\%$ of pairs produce inconsistent verdicts across runs under Vanilla RAG; Ours-RG cuts this to $4.9\%$ ($-52\%$), with high-uncertainty pairs (majority verdict $<$80\%) dropping from 58 to 28 ($-52\%$). Sonnet at $T=0.7$ shows the same direction (VR $7.8\%$ inconsistent, $48$ high-uncertainty $\to$ Ours-RG $3.4\%$, $21$). The two Haiku $T=0.7$ Ours-Full runs with anomalously long evaluation vectors ($n>50$) are filtered before aggregation to avoid spuriously deflating entropy via single-verdict zero-entropy padding.

\paragraph{Difficulty Stratification (Per-Probe Counts).} On Haiku ($n=38$ hard probes), Ours-Full drives 31 probes at $T=0$, 24 at $T=0.5$, and 24 at $T=0.7$ to perfect verdict consistency (VVIR$=1.0$); on Sonnet ($n=23$ hard probes), 22 / 15 / 13 respectively. These per-temperature counts complement the sign-test result reported in Section~\ref{sec:determinism}.

\paragraph{Memory-baseline comparison.} Both query-centric memory baselines (Reflexion, ReasoningBank) also reduce hard-stratum VVIR significantly at $T{=}0$ (Table~\ref{tab:vvir_baselines}), contradicting any conceptual prediction that variance reduction is exclusive to evidence-centric memory. Quantitatively, however, evidence-centric memory captures a substantially larger fraction of the available headroom ($1{-}\text{VR}_{\text{VVIR}}$): Ours-Full reaches $87$--$97\%$ vs.\ $33$--$48\%$ for the query-centric baselines. Ours-Full's Sonnet hard-stratum VVIR (0.987) is statistically indistinguishable from perfect at $K{=}10$. The baseline plateau is consistent across both models and persists at higher temperatures: at $T{=}0.7$, ReasoningBank reaches 0.692/0.803 hard-stratum VVIR on Haiku/Sonnet while Ours-Full holds at 0.881/0.927.

\begin{table}[t]
\centering
\caption{Hard-stratum VVIR at $T{=}0$ across all memory baselines on the paired determinism probe set (Haiku $n{=}38$, Sonnet $n{=}23$, $K{=}10$ runs each). Headroom captured $= \Delta / (1{-}\text{VR}_{\text{VVIR}})$. Both query-centric baselines (Reflexion, ReasoningBank) significantly reduce VVIR, but evidence-centric memory captures roughly twice as much of the available headroom on Haiku and roughly $2\times$ on Sonnet. $p$-values are paired one-sided Wilcoxon signed-rank vs.\ Vanilla RAG. Ours-RG and Ours-Full $p<10^{-4}$ throughout (omitted).}
\label{tab:vvir_baselines}
\small
\begin{tabular}{lcccc}
\toprule
\textbf{Method} & \textbf{Haiku VVIR} & \textbf{Headroom} & \textbf{Sonnet VVIR} & \textbf{Headroom} \\
\midrule
Vanilla RAG       & 0.650            & --   & 0.557            & --   \\
Reflexion         & 0.766 ($p{=}0.006$)        & 33\% & 0.748 ($p{=}5{\times}10^{-4}$) & 43\% \\
ReasoningBank     & 0.816 ($p{=}5{\times}10^{-4}$) & 47\% & 0.770 ($p{=}8{\times}10^{-4}$) & 48\% \\
Ours-RG           & 0.863                       & 61\% & 0.883                          & 74\% \\
\textbf{Ours-Full}& \textbf{0.953}              & \textbf{87\%} & \textbf{0.987}        & \textbf{97\%} \\
\bottomrule
\end{tabular}
\end{table}

\subsection{High-Reuse Efficiency and Cost Drilldown}
\label{sec:enh_efficiency_drill}

\paragraph{Efficiency and Pruning.} Candidates are pruned from 20 to 5.8 by the end of the 387-question high-reuse run (70.9\% steady-state reduction; 57\% averaged over the full run), driving 40\% total token savings. Ours-Full vs.\ Ours-RG accuracy difference is not significant on either model ($p>0.07$; Sonnet $p=0.073$, 25:13; Haiku $p=0.49$, 19:14), confirming that the retrieval graph's role in high-reuse is primarily efficiency: comparable accuracy but 45\% lower latency on Sonnet ($8.46$s vs.\ $15.31$s). Speed gains are reuse-density-dependent; on standard MuSiQue (lighter pruning: 20$\to 16.1$ candidates at the per-cluster steady state), Ours-Full is slower ($42.3$s vs.\ $15.8$s for Vanilla RAG) because graph traversal overhead dominates. The efficiency story requires sufficient passage reuse for aggressive pruning to offset overhead.

\paragraph{Cost and Latency.} In high-reuse settings, Ours-Full is 46\% cheaper and 46\% faster than Vanilla RAG, achieving 94 correct answers per dollar vs.\ 45 for Haiku Vanilla RAG ($\sim$$2\times$ more efficient). In standard MuSiQue, Ours-Full is $2.7\times$ slower because graph traversal overhead dominates with lighter pruning ($20\to 16.1$ candidates at the per-cluster steady state); per-query token cost is nonetheless slightly lower than Vanilla RAG ($2.65$\textcent{} vs.\ $2.89$\textcent, $-8\%$), as tighter per-evaluation contexts under partial pruning offset the additional traversal calls. Efficiency is directly tied to passage reuse density. Cross-model transfer of evidence profiles (initializing a smaller model with a larger model's reasoning graph) is left to future work.

\subsection{Multiple-Comparisons Adjusted $p$-Values}
\label{app:multiplicity}

Across the family of approximately 30 reported $p$-values in the main paper, we apply both Holm step-down correction (within Tab~\ref{tab:main}'s 4-contrast family) and Benjamini-Hochberg false discovery rate (BH-FDR) control across the full $\sim$30-test family at $\alpha=0.05$.

\paragraph{Headline contrasts surviving any reasonable correction.}
\begin{itemize}
    \item Ours-Full vs.\ Vanilla RAG combined paired ($N=1{,}600$): raw $p=1.9\times10^{-5}$; Holm $p_{\text{adj}}=7.6\times10^{-5}$; BH $q\approx 5.7\times10^{-4}$. Survives.
    \item 50\%+ coverage stratum, Ours-Full vs.\ Vanilla RAG ($n=263$): raw $p=1.9\times10^{-6}$; BH $q\approx 5.7\times10^{-5}$. Survives.
    \item Sonnet hard-stratum VVIR at $T=0$, Ours-Full vs.\ Vanilla RAG: raw $p<10^{-4}$; BH $q<3\times10^{-3}$. Survives.
    \item 4-hop F1, Ours-Full vs.\ Vanilla RAG: raw $p=0.0001$; BH $q\approx 3\times10^{-3}$. Survives.
\end{itemize}

\paragraph{Marginal contrasts that do not survive (already labeled n.s.\ in body).}
\begin{itemize}
    \item Ours-Full vs.\ ReasoningBank, 50\%+ coverage stratum: raw $p=0.064$; BH $q\approx 0.10$. Does not survive at $\alpha=0.05$ (already labeled non-significant in body).
    \item Same comparison at 60\%+ ($p=0.078$) and 70\%+ ($p=0.18$) thresholds: do not survive (already labeled non-significant).
\end{itemize}

The four headline contrasts that survive correction support all numerical claims in the abstract; the marginal ReasoningBank contrasts that do not survive are already disclosed as non-significant in Section~\ref{sec:dose_response}, so multiple-comparisons correction does not retract any claim made in the paper.

\subsection{Method Supplements: Running Example, Convergence, Ceiling}
\label{sec:enh_method_supp}

This appendix subsection collects illustrative and analytical material from Section~\ref{sec:feedback_loop} that complements the formal mechanism in Sections~\ref{sec:reasoning_graphs}--\ref{sec:retrieval_graphs}.

\paragraph{Running Example.} Consider a cluster of multi-hop questions about award-winning films (\eg, ``Who directed the film that won the Palme d'Or in 2019?'', requiring retrieval of passages about the award ceremony and the winning director):
\begin{itemize}
    \item \textbf{Run 1}: The agent retrieves 12 candidate passages and reasons from scratch. A passage about a similarly-titled but different film ($k_{37}$) is incorrectly marked \texttt{used}, but the agent arrives at the correct answer via other evidence. Evaluated edges and a retrieved\_via edge are written.
    \item \textbf{Run 2}: A related question is posed. The pipeline planner excludes 3 previously-rejected passages (funnel: $12 \rightarrow 9$). For each remaining passage, the system injects an evidence profile. The agent sees: ``$k_{37}$ was rejected in 1 prior correct decision because: describes a different film with a similar title, different year.'' The agent avoids the same confusion.
    \item \textbf{Run $N$}: The funnel narrows to 7 passages. Evidence profiles are rich. The agent consistently identifies the correct director without being misled by similarly-titled passages.
\end{itemize}

\noindent To illustrate concretely, consider passage $k_{37}$, which contains: \emph{``The 2018 Palme d'Or was awarded to Shoplifters, directed by Hirokazu Kore-eda\ldots''}. On Run~2, the agent receives this passage alongside the following injected profile:

\begin{quote}
\small
\texttt{[EVIDENCE PROFILE] Evaluated 1 time in prior correct decisions.} \\
\texttt{\ \ Verdict distribution: used 0/1, rejected 1/1.} \\
\texttt{\ \ Reliability score: 0.00} \\
\texttt{\ \ Top reason for ``rejected'': ``describes a different film with a similar title, different year''}
\end{quote}

\noindent \textbf{Without profile} (Run~1): the agent marks $k_{37}$ as \texttt{used}, confusing the 2018 winner with the 2019 query. \textbf{With profile} (Run~2): the agent sees that $k_{37}$ was rejected in every prior correct decision for describing the wrong year, and correctly rejects it. By Run~$N$, after 28 evaluations, the profile shows rejected 27/28, reliability 0.04, and the verdict is stable.

\subsection{Additional Limitations}
\label{sec:enh_limitations_more}

The body Section~\ref{sec:limitations} covers the five limitations most directly tied to the deployment claim (dataset scope, reuse-dependent efficiency, aggressive-pruning trade-off, cold start, error reinforcement). The remaining limitations, with full discussion, are:

\textbf{Model family.} Evaluation covers two closed-source families (Anthropic Claude Haiku~4.5 / Sonnet~4 and OpenAI GPT-5-mini; cross-vendor replication in Appendix~\ref{sec:enh_crossvendor}). Generalization to open-source models (Llama, Qwen, Mistral) and to future architectures is untested and remains an open empirical question.

\textbf{Determinism experiment scale.} VVIR is measured on a paired $N=73$ probe set ($K=10$, three temperatures, both Claude Haiku 4.5 and Claude Sonnet 4); the canonical 30-probe Haiku subset is used for the cp50/cp200/cp387 convergence diagnostic. Absolute deltas may shift on a different MuSiQue subsample, but cross-model and cross-temperature direction is consistent (Table~\ref{tab:vvir}; per-subset sensitivity in Appendix~\ref{app:probe_sensitivity}).

\textbf{Cross-model transfer.} Cross-model transfer (initializing one model with another's reasoning graph) is unevaluated in this work and left to future investigation.

\textbf{Sequential protocol.} Questions are processed sequentially within clusters; ordering robustness verified across two additional seeds (Appendix~\ref{app:ordering}). Real deployments may have different arrival rates and concurrency patterns.

\textbf{Context window limits.} Evidence profiles consume tokens. The selection policy (Section~\ref{sec:reasoning_graphs}) bounds total injected context, but aggressive truncation may lose useful signal.

\textbf{Profile granularity.} Current profiles aggregate all past evaluations of an evidence item regardless of query context or recency. Structured filtering by query type, temporal weighting toward recent evaluations, or embedding-based profile matching could improve precision but are not yet implemented.

\subsection{Conditional-Dilution Forest Plot, Hop-Depth Table, Verdict-Grid Visualization}
\label{sec:enh_visuals}

The body Section~\ref{sec:results} reports the conditional-dilution result, the hop-depth scaling, and the verdict-grid example as compressed prose; the original visuals/tables are reproduced here. The reasoning-graph anatomy figure (forward/backward traversal) referenced from Section~\ref{sec:reasoning_graphs} is also reproduced.

\paragraph{Verdict vs.\ answer determinism (Sec.~\ref{sec:determinism}, expanded).} On Ours-RG perfect-verdict probes at $T{=}0$, Haiku has 47 (14 correct-consistent / 15 wrong-consistent / 18 verdict-identical-but-answer-varying) and Sonnet 58 (28/22/8). The wrong-consistent residue is structurally distinct from the consistency-amplifies-wrong-interpretation failure mode: the correct-outcome filter (Section~\ref{sec:reasoning_graphs}) blocks profiles from carrying forward verdicts that historically led to wrong answers, so the residue lives downstream of verdict alignment, in answer-synthesis. The VVIR claim is therefore scoped to evidence-evaluation consistency, not end-to-end answer correctness.

\paragraph{When to use evidence profiles (decision rules, Sec.~\ref{sec:limitations}).} Largest benefit when (i) gold-passage reuse $>$$\sim$0.10 (Jaccard is unreliable as a proxy: 2Wiki Jaccard 0.499/gold 0.025 is null vs.\ MuSiQue 0.504/0.441 $+3.55$pp); (ii) $\geq$50 questions per cluster type so profiles densify; (iii) multi-hop reasoning ($+11$pp 4-hop vs.\ $+1.7$pp 2-hop); and (iv) audit/safety-critical use, where the cross-model $+8$--$13$pp confound-free VVIR lift on Ours-RG dominates: prefer Ours-RG (no pruning) or Ours-Full with conservative $R_{\text{thresh}}$.

\begin{figure}[h]
    \centering
    \begin{tikzpicture}[
        agent/.style={circle, draw, minimum size=0.8cm, font=\scriptsize\bfseries},
        agentfade/.style={circle, draw, minimum size=0.55cm, font=\tiny, opacity=0.35},
        decision/.style={draw, rounded corners=2pt, minimum height=0.6cm, minimum width=1.3cm, font=\scriptsize, align=center},
        passage/.style={draw, rounded corners=2pt, minimum height=0.6cm, minimum width=1.0cm, font=\scriptsize, align=center},
        passagerej/.style={passage, fill=black!8},
        arr/.style={-{Stealth[length=4pt]}, semithick},
        arrleft/.style={{Stealth[length=4pt]}-, semithick},
        elabel/.style={font=\tiny, align=center},
    ]
    \node[font=\small\bfseries] (fwd) at (0, 0) {Forward Traversal};
    \node[agent, below=0.8cm of fwd] (ag) {$a$};
    \node[decision, right=2.5cm of ag, yshift=-0.5cm] (dec) {$d_{42}$};
    \draw[arr] (ag) -- node[elabel, above] {decided} (dec);
    \node[elabel, below=0.15cm of dec] {\texttt{conf: 0.91, correct}};
    \node[passage, below=2.0cm of ag, xshift=-2.0cm] (k12) {$k_{12}$};
    \node[passagerej, below=2.0cm of ag] (k37) {$k_{37}$};
    \node[passage, below=2.0cm of ag, xshift=2.0cm] (k8) {$k_{8}$};
    \draw[arr] (ag) -- node[elabel, left] {\texttt{used}} (k12);
    \draw[arr] (ag) -- node[elabel, right] {\texttt{rejected}} (k37);
    \draw[arr] (ag) -- node[elabel, right] {\texttt{used}} (k8);
    \draw[dotted, thick] (5.2, 0.3) -- (5.2, -5.0);
    \node[font=\small\bfseries] (bwd) at (9.0, 0) {Backward Traversal};
    \node[passagerej, below=0.8cm of bwd, minimum width=1.3cm] (k37b) {$k_{37}$};
    \node[agentfade, above left=1.0cm and 0.9cm of k37b] (a1) {$a_1$};
    \node[agentfade, left=1.1cm of k37b] (a2) {$a_2$};
    \node[agentfade, below left=1.0cm and 0.9cm of k37b] (a3) {$a_3$};
    \node[agentfade, above right=1.0cm and 0.9cm of k37b] (a13) {$a_{13}$};
    \node[agentfade, right=1.1cm of k37b] (a14) {$a_{14}$};
    \node[font=\tiny, opacity=0.35, below right=1.0cm and 0.9cm of k37b] (dots) {$\cdots$};
    \draw[arrleft, opacity=0.35] (a1) -- (k37b);
    \draw[arrleft, opacity=0.35] (a2) -- (k37b);
    \draw[arrleft, opacity=0.35] (a3) -- (k37b);
    \draw[arrleft, opacity=0.35] (a13) -- (k37b);
    \draw[arrleft, opacity=0.35] (a14) -- (k37b);
    \node[draw, rounded corners=3pt, below=1.8cm of k37b, font=\scriptsize, align=left, text width=4.8cm] (profile) {%
    \textbf{Evidence Profile}\\[2pt]
    Verdicts: \texttt{rejected} 13/14, \texttt{used} 1/14\\
    Top reason: ``different film, similar title''\\
    Reliability: $R(k_{37}, \tau) = 0.07$};
    \draw[arr, densely dashed] (k37b) -- (profile);
    \end{tikzpicture}
    \caption{Reasoning graph anatomy. \emph{Forward} (left): from an agent through decided/evaluated edges, reconstruct one run's chain of thought. \emph{Backward} (right): from an evidence item inward, aggregate evaluations across runs to surface cross-run patterns. Backward traversal enables evidence-centric feedback; flat strategy stores cannot support it.}
    \label{fig:anatomy}
\end{figure}
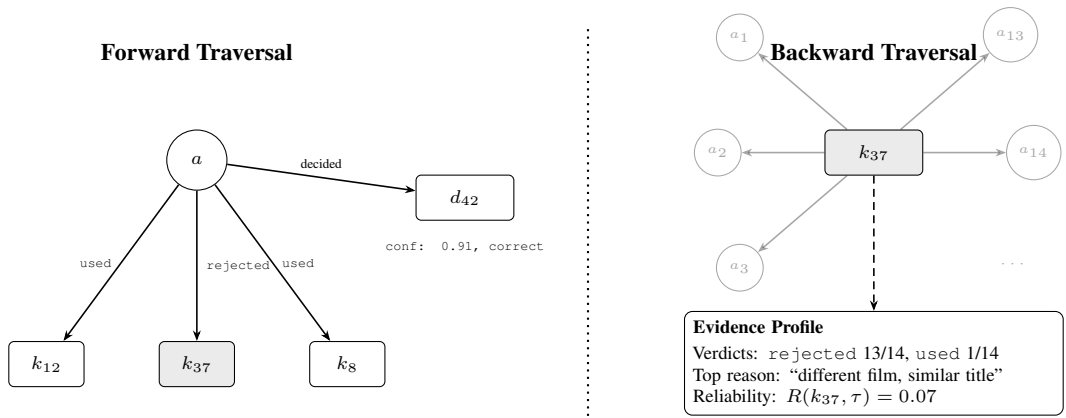

\begin{figure}[h]
    \centering
    \includegraphics[width=0.85\textwidth]{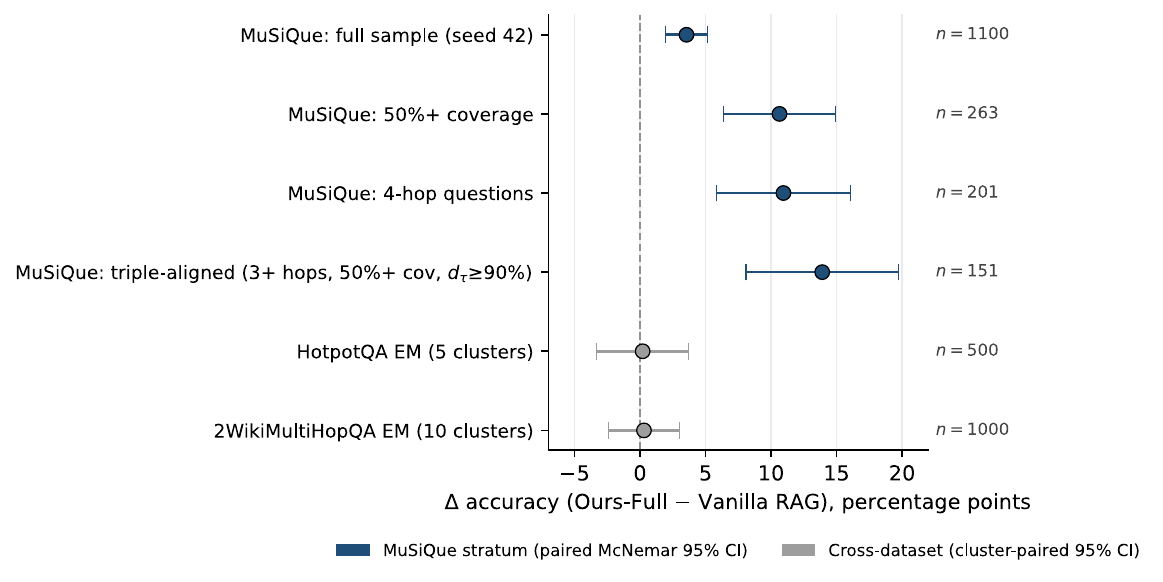}
    \caption{Conditional dilution of the headline effect. Each row shows $\Delta$ accuracy (Ours-Full $-$ Vanilla RAG) for one stratum or dataset, with paired 95\% confidence intervals. MuSiQue strata use McNemar discordant-pair CIs; cross-dataset rows use cluster-paired CIs from Table~\ref{tab:main}. The full-sample MuSiQue effect ($+3.55$pp) is a $4\times$ underestimate of the conditional effect on triple-aligned questions ($+13.9$pp). HotpotQA and 2Wiki sit at zero, consistent with the gold-passage reuse mechanism.}
    \label{fig:strata}
\end{figure}

\begin{table}[h]
\centering
\caption{Accuracy by reasoning depth. Improvement scales monotonically with hop count.}
\label{tab:hop_depth}
\small
\begin{tabular}{lccccl}
\toprule
\textbf{Hops} & \textbf{$n$} & \textbf{Ours-Full} & \textbf{VR} & \textbf{$\Delta$} & \textbf{McNemar} \\
\midrule
2-hop & 646 & 68.9\% & 67.2\% & +1.7pp & n.s. \\
3-hop & 253 & 66.0\% & 63.6\% & +2.4pp & n.s. \\
\textbf{4-hop} & \textbf{201} & \textbf{68.2\%} & \textbf{57.2\%} & \textbf{+11.0pp} & $\boldsymbol{p=0.0001}$ \\
\bottomrule
\end{tabular}
\end{table}

\begin{figure}[h]
    \centering
    \includegraphics[width=0.7\textwidth]{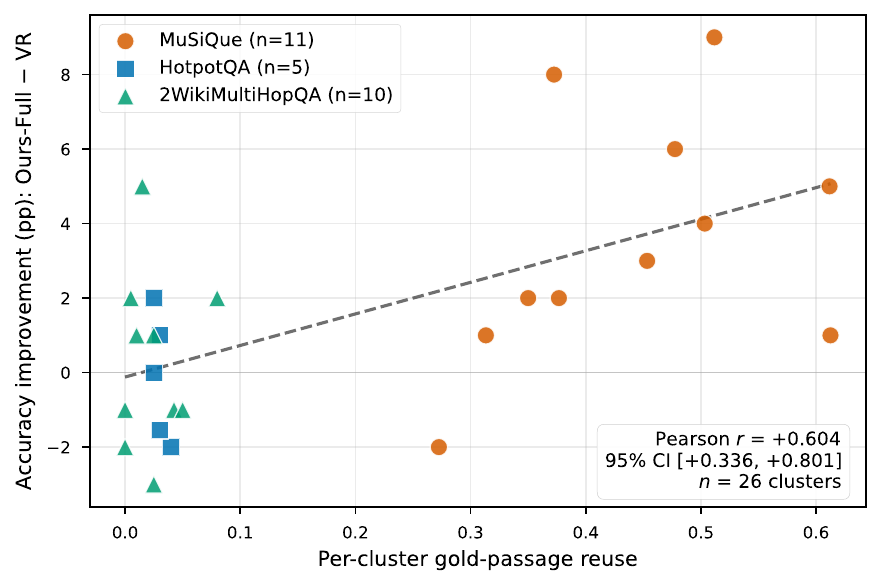}
    \caption{Per-cluster gold-passage reuse vs.\ accuracy improvement (Ours-Full $-$ Vanilla RAG); $n=26$ clusters across MuSiQue, HotpotQA, 2WikiMultiHopQA. High-reuse MuSiQue clusters show substantial improvements; near-zero-reuse HotpotQA and 2Wiki clusters cluster near zero. Pearson $r=0.604$, 95\% CI $[0.336, 0.801]$ (10{,}000-resample bootstrap). Cited in Section~\ref{sec:hop_depth}.}
    \label{fig:overlap_scatter}
\end{figure}

\paragraph{Empirical Convergence.} We observe that $\atau(n)$ improves over successive runs as profiles densify. Two mechanisms drive this trend: the retrieval graph gradually excludes evidence items with high rejection rates, tightening the candidate funnel; and the reasoning graph increases the evaluation context available per evidence item, reducing the probability of incorrect verdicts. Both effects compound when outcome signals are reliable and query types are stable, conditions that hold in our experimental setting and that we validate empirically in Section~\ref{sec:results}. We do not claim formal convergence guarantees; the trend is empirically demonstrated through the dose-response analysis (Section~\ref{sec:dose_response}).

\paragraph{Why the Ceiling Is High.} Define evidence profile coverage $\pcov(n)$ as the fraction of items in $C_t$ with at least one prior evaluation after $n$ runs. As $n$ grows, $\pcov(n)$ increases toward 1 because the same items recur across queries of the same type. Once $\pcov(n) \approx 1$, the agent never reasons from scratch on any evidence item. The remaining error rate is dominated by three sources: (a) novel items ($\pcov < 1$), which shrinks over time; (b) misleading evaluation histories, which are rare because profiles aggregate across many runs; and (c) irreducible query ambiguity, the task-specific floor. For well-defined tasks, source (c) is small, and sources (a) and (b) shrink as the graph densifies.

\section{Algorithm Pseudocode}
\label{app:algorithms}

\begin{algorithm}[h]
\caption{Evidence-Centric Context Injection}
\label{alg:injection}
\begin{algorithmic}[1]
\REQUIRE Candidate set $C_t$, Reasoning graph $\Gcal_R$, Token budget $B$, Sampling threshold $T_{\max}$, Sample size $N$
\ENSURE Augmented prompt with evidence profiles
\STATE $\text{profiles} \leftarrow \{\}$
\FOR{each $k_i \in C_t$}
    \STATE $\Ein(k_i) \leftarrow \{e \in \Eeval \mid e = (*, k_i, *)\}$
    \STATE Filter $\Ein(k_i)$ for edges with $\text{outcome} = \texttt{correct}$
    \IF{$|\Ein(k_i)| = 0$}
        \STATE \textbf{continue} \COMMENT{Cold start: agent reasons from scratch for $k_i$}
    \ENDIF
    \IF{$|\Ein(k_i)| > T_{\max}$}
        \STATE Sample $N$ most recent correct evaluations
    \ENDIF
    \STATE Compute verdict distribution, representative reasons, $R(k_i, \tau)$
    \STATE $\text{profiles}[k_i] \leftarrow \mathcal{P}(k_i)$
\ENDFOR
\STATE Rank profiles by $|\Ein(k_i)|$ descending
\STATE Truncate to fit within token budget $B$
\STATE Inject profiles into prompt alongside raw evidence
\RETURN augmented prompt
\end{algorithmic}
\end{algorithm}

\begin{algorithm}[h]
\caption{Pipeline Planner (Retrieval Graph Optimization)}
\label{alg:planner}
\begin{algorithmic}[1]
\REQUIRE Query type $\tau$, Retrieval graph $\Gcal_P$, Reasoning graph $\Gcal_R$, Min support, $R_{\text{thresh}}$
\ENSURE Optimized filter configuration $f^*$, Exclusion list
\STATE Collect all retrieved\_via edges for queries of type $\tau$
\FOR{each distinct filter configuration $f$}
    \STATE $S(f) \leftarrow |\{e : e.\text{filter} = f \wedge \text{outcome} = \texttt{correct}\}| / |\{e : e.\text{filter} = f\}|$
\ENDFOR
\STATE $f^* \leftarrow \arg\max_f S(f)$ s.t.\ support $\geq$ min\_support
\STATE Query $\Gcal_R$ for items with rejection rate $> R_{\text{thresh}}$ for type $\tau$
\STATE Add high-rejection items to exclusion list
\RETURN $f^*$, exclusion list
\end{algorithmic}
\end{algorithm}

\begin{algorithm}[h]
\caption{Edge Write Protocol}
\label{alg:write}
\begin{algorithmic}[1]
\REQUIRE Agent $a$, Query $q_t$, Candidate set $C_t$, Action $a_t$, Structured CoT $\theta_t$
\ENSURE Graphs updated atomically
\STATE Create decision node $d_t$
\STATE Write decided edge: $(a, d_t, \phi_t)$ with $\phi_t = (\text{confidence}, \text{timestamp}, \text{pending})$
\FOR{each $(k_i, v_i, r_i, \delta_i) \in \theta_t$}
    \STATE Write evaluated edge: $(a, k_i, \psi_i)$ with $\psi_i = (\text{step}, v_i, r_i, \delta_i, d_t)$
\ENDFOR
\STATE Write retrieved\_via edge: $(d_t, q_t, \rho_t)$ with pipeline metadata
\STATE \COMMENT{Outcome field on decided edge populated when signal arrives}
\end{algorithmic}
\end{algorithm}

\section{Hyperparameter Table}
\label{app:hyperparams}

\begin{table}[h]
\centering
\caption{Hyperparameters and their values.}
\label{tab:hyperparams}
\small
\begin{tabular}{llp{7cm}}
\toprule
\textbf{Symbol} & \textbf{Value} & \textbf{Description} \\
\midrule
$T_{\max}$ & 50 & Max evaluations per evidence item before sampling \\
$N$ & 20 & Number of recent evaluations to sample when $|\Ein| > T_{\max}$ \\
$B$ & 2000 & Token budget for injected evidence profiles \\
$R_{\text{thresh}}$ & 0.7 & Rejection rate threshold for exclusion list \\
min\_support & 3 & Minimum observations before optimizing filter \\
$k$ (MuSiQue) & 25 & Number of clusters for query type classification \\
$|C_t|$ & 20 / 10 & Initial candidate set size (top-$k$ retrieval); 20 for MuSiQue and high-reuse, 10 for HotpotQA and 2WikiMultiHopQA \\
Temperature & 0 & LLM sampling temperature (0.5 and 0.7 for determinism robustness tests) \\
\bottomrule
\end{tabular}
\end{table}

\section{Clustering Sensitivity}
\label{app:clustering}

We cluster questions using BAAI/bge-base-en-v1.5 embeddings with $k$-means ($k{=}25$ for MuSiQue, $k{=}15$ per type for HotpotQA, $k{=}12$ globally for 2WikiMultiHopQA). Clusters smaller than a minimum size (50 for MuSiQue, 100 for HotpotQA, 80 for 2WikiMultiHopQA) are merged into the nearest cluster by centroid cosine distance. MuSiQue produces 21 clusters after merging (sizes 56--235); HotpotQA produces 21 clusters (15 bridge, 6 comparison). For the evaluation, we select the 11 largest MuSiQue clusters and sample 100 questions from each (1,100 total). For HotpotQA, we select 5 clusters (1 bridge, 4 comparison) with 100 questions each (500 total). For 2WikiMultiHopQA, we use 10 clusters (after merging clusters below the 80-question threshold from an initial $k{=}12$) with 100 questions each (1{,}000 total); cluster identifiers are prefixed \texttt{twowiki\_}. All clustering uses random seed 42 and scikit-learn's \texttt{KMeans} with \texttt{n\_init=10}.

The per-cluster analysis in Section~\ref{sec:analysis} shows 10/11 MuSiQue clusters benefiting (paired $t$-test $p=0.0049$, Wilcoxon $p=0.0068$). The single negative cluster (musique\_2, $-$2.0pp) has the second-lowest gold-passage reuse within MuSiQue (0.273, vs.\ cluster mean 0.441), consistent with the pooled gold-reuse correlation ($r=0.604$, $p=0.001$, $n=26$; Section~\ref{sec:hop_depth}). This suggests the results are robust to specific cluster assignments: improvement is driven by gold-passage reuse density, not by favorable clustering.

\section{Storage Growth and Pruning}
\label{app:pruning}

The reasoning graph grows linearly with the number of queries processed. Each query produces one decided edge and $|C_t|$ evaluated edges (one per candidate passage). With $|C_t|{=}20$, this is 21 edges per query. Over our experiments:

\begin{itemize}
    \item \textbf{MuSiQue} (1,100 queries): $\sim$23,100 edges (1,100 decided + 22,000 evaluated).
    \item \textbf{High-Reuse} (387 queries): $\sim$8,127 edges before pruning. With retrieval graph pruning, $|C_t|$ drops from 20 to 5.8 at steady state (last-50 mean), reducing the steady-state edge write rate by 70.9\%.
    \item \textbf{Retrieval graph}: 1 retrieved\_via edge per query. Negligible relative to evaluated edges.
\end{itemize}

Graphs are stored as JSON files. The largest graph (MuSiQue Ours-Full, 1,100 queries) occupies $\sim$12\,MB on disk. Graph reads (backward traversal for evidence profiles) complete in $<$50\,ms for all experiments, well within the latency budget dominated by LLM inference ($\sim$3--15\,s per query).

For production workloads (millions of queries), a graph database backend (e.g., SurrealDB, Neo4j) would replace JSON storage. Edge pruning strategies (such as retaining only the $N$ most recent evaluations per passage or decaying edge weights over time) would bound storage growth. Our selection policy (Section~\ref{sec:reasoning_graphs}) already bounds \emph{read} cost by sampling when $|\Ein(k_i)| > T_{\max}$; a complementary \emph{write} pruning policy would bound storage. We leave empirical evaluation of pruning strategies for future work.

\section{Prompt Templates}
\label{app:prompts}

The agent uses tool calling to produce structured per-evidence reasoning. The system message and tool schema are shown below. When evidence profiles are available, they are injected inline after each passage.

\paragraph{System Message.}
\begin{quote}
\small
\texttt{You are a multi-hop question answering agent. You will receive a question and a set of passages. Your task is to:}

\texttt{1. Evaluate EVERY passage: determine whether it is relevant ("used") or irrelevant ("rejected") to answering the question. You must provide a verdict for every passage.}

\texttt{2. Provide a brief reason for each verdict.}

\texttt{3. Provide a confidence delta (-1 to 1) indicating how much each passage shifts your confidence in the answer.}

\texttt{4. Provide your final answer to the question.}

\texttt{Some passages may include prior evaluation history from previous runs. This history shows how the passage has been judged before; use it to inform your evaluation, but apply your own judgment for the current question.}
\end{quote}

\paragraph{Tool Schema.} The model is forced to call a \texttt{submit\_answer} tool with two required fields: (1) \texttt{evidence\_evaluations}, an array of objects each containing \texttt{passage\_id} (string), \texttt{verdict} (enum: \texttt{used}/\texttt{rejected}), \texttt{reason} (string), and \texttt{confidence\_delta} (number $\in [-1, 1]$); and (2) \texttt{final\_answer} (string).

\paragraph{Evidence Profile Format.} Profiles are injected immediately after each passage's text:

\begin{quote}
\small
\texttt{[EVIDENCE PROFILE] Evaluated 28 times in prior correct decisions.} \\
\texttt{\ \ Verdict distribution: used 1/28, rejected 27/28.} \\
\texttt{\ \ Reliability score: 0.04} \\
\texttt{\ \ Top reason for "rejected": "describes a different film with a similar title"}
\end{quote}

\paragraph{ReasoningBank Distillation Prompt.} After each query, the ReasoningBank baseline calls the model with a prompt containing the question, answer, outcome, and per-passage evaluations, and asks it to distill a reusable strategy triple \texttt{\{title, description, content\}} in JSON format. Strategies are embedded using BAAI/bge-base-en-v1.5 and the top-3 most similar strategies are injected into future queries.

\paragraph{Reflexion Prompt.} After each query, the Reflexion baseline appends a self-reflection to a text buffer. The 3 most recent reflections are injected at the top of the prompt for subsequent queries.

\section{Ordering Robustness}
\label{app:ordering}

To verify that convergence is not an artifact of the specific question order, we re-ran the high-reuse experiment (387 questions) with two additional ordering seeds (43, 44) for both Ours-Full and Vanilla RAG on Haiku. All six runs completed the full 387 questions.

\begin{table}[h]
\centering
\caption{Ordering robustness: Ours-Full and Vanilla RAG accuracy across three ordering seeds on the high-reuse dataset (Haiku). Ours-Full converges to a stable accuracy regardless of question order (range 56.6--58.7\%, std 0.86pp at $n{=}387$). The delta vs.\ Vanilla RAG at position 200 is consistently positive across all orderings.}
\label{tab:ordering}
\small
\begin{tabular}{lcccc}
\toprule
\textbf{Seed} & \textbf{Ours-Full} & \textbf{VR @200} & \textbf{$\Delta$ @200} & \textbf{Ours-Full F1} \\
\midrule
42 & 56.6\% ($n{=}387$) & 51.0\% & +5.6pp  & 0.633 \\
43 & 57.4\% ($n{=}387$) & 46.5\% & +10.9pp & 0.632 \\
44 & 58.7\% ($n{=}387$) & 53.0\% & +5.7pp  & 0.638 \\
\midrule
Mean & 57.5\% (std 0.86pp) & 50.2\% (std 2.74pp) & +7.4pp & 0.634 (std 0.003) \\
\bottomrule
\end{tabular}
\end{table}

The key finding is that Ours-Full accuracy is order-stable: 56.6--58.7\% across three orderings (2.1pp spread, std 0.86pp). Mean token F1 is similarly stable (0.632--0.638, std 0.003). Early in the sequence, ordering introduces variability, but this narrows as profiles accumulate, reaching std 0.86pp at convergence. This confirms that the feedback loop converges to a comparable equilibrium regardless of the order in which profiles are built.

Vanilla RAG shows higher variability across orderings at position 200 (std 2.74pp), but this reflects which questions happen to fall in the first 200 under each ordering, not an ordering effect; Vanilla RAG has no memory.

\paragraph{MuSiQue Multi-Seed Diagnostic.}
The Table~\ref{tab:main} headline (+3.55pp, seed 42) is one of three Sonnet ordering seeds for MuSiQue $\{42, 7, 13\}$, with per-seed deltas $+3.55 / +1.18 / +2.64$ pp (per-seed McNemar $p = 1.7\!\times\!10^{-5}$, $0.165$, $3.1\!\times\!10^{-3}$). Seed~7's attenuation merits explanation. We performed a paired diagnostic across all 11 MuSiQue clusters and find the underperformance is mechanism-specific and stratified, not a uniform ordering hardness:
\begin{itemize}
\item \textbf{Vanilla RAG accuracy is essentially flat across seeds} (64.55\% / 64.91\% / 66.36\% on seeds 42/7/13). The protocol is not intrinsically harder under seed~7; only Ours-Full loses traction.
\item \textbf{Coverage distributions are essentially identical across seeds} (mean per-question coverage 0.250 / 0.236 / 0.242; fraction of questions at coverage~$>0.50$: 21.8\% / 20.0\% / 20.0\%). Ours-Full builds approximately the same volume of evidence on seed~7.
\item \textbf{The high-coverage dose-response is preserved on every seed.} At coverage~$\in (0.50, 0.75]$, Ours-Full vs.\ Vanilla RAG paired deltas are $+8.2 / +6.3 / +4.5$pp on seeds 42/7/13. At coverage~$> 0.75$, deltas are $+13.8 / +6.4 / +12.8$pp. The conditional claim that drives the paper's headline empirical contribution survives the most adversarial of the three seeds.
\item \textbf{Seed~7's attenuation is concentrated in the mid-coverage stratum.} At coverage~$\in (0.25, 0.50]$, Ours-Full underperforms Vanilla RAG by $-1.8$pp under seed~7 versus $+5.8 / +8.4$pp under seeds 42/13. The deficit is driven by 2--3 specific clusters (notably 1, 2, 14) where seed~7's ordering places hard early questions before sufficient corrective evidence has accumulated, producing transient mid-coverage misdirection that recovers as more queries are processed.
\end{itemize}
The pattern is consistent with order-induced early misdirection that is corrected as evidence accumulates: profile signal in the mid-coverage range can amplify a few unlucky early evaluations until enough additional signal arrives to dilute them. The load-bearing high-coverage conditional claim is preserved on every seed including the adversarial one, while the average-case effect varies more substantially with ordering than the conditional effect does. This is the expected behavior of a mechanism whose contribution is concentrated in a stratum of the input distribution.

\section{Outcome Signal Noise Robustness}
\label{app:noise}

A concern with any feedback loop is sensitivity to noisy outcome signals. In deployment, the ``correct'' / ``incorrect'' labels used to build evidence profiles may come from unreliable downstream signals rather than gold labels. To test robustness, we re-ran the full sequential protocol (Ours-Full, Haiku, high-reuse dataset, 387 questions) with artificially corrupted outcome labels at three noise levels: 10\%, 20\%, and 30\%. At each level, the specified fraction of outcomes are randomly flipped \emph{before} being written to the graph (so evidence profiles accumulate noisy data) while the true outcome is preserved for accuracy measurement. At 10\% noise, we ran three independent noise seeds to quantify variance from the stochastic flip pattern.

\begin{table}[h]
\centering
\caption{Noise robustness: Ours-Full accuracy under outcome label noise on the high-reuse dataset (Haiku). The 10\% row reports mean $\pm$ std across 3 noise seeds. Even at 30\% label noise, the system remains above the current Vanilla RAG baseline (49.1\%, Table~\ref{tab:highreuse}).}
\label{tab:noise}
\small
\begin{tabular}{lcccc}
\toprule
\textbf{Noise Rate} & \textbf{Accuracy} & \textbf{$\Delta$ vs.\ VR} & \textbf{$\Delta$ vs.\ Clean} & \textbf{Flips} \\
\midrule
0\% (clean) & 56.6\% & +7.5pp & n/a & 0 \\
10\% (3 seeds) & 56.4\% $\pm$ 3.5pp & +7.3pp & $-$0.2pp & 42 \\
20\% & 52.5\% & +3.4pp & $-$4.1pp & 71 \\
30\% & 52.7\% & +3.6pp & $-$3.9pp & 118 \\
\bottomrule
\end{tabular}
\end{table}

At 10\% noise, accuracy across three noise seeds averages 56.4\% (std 3.5pp, range 53.0--59.9\%), compared to 56.6\% clean, a negligible $-$0.2pp mean difference. The high seed-to-seed variance (3.5pp) indicates that which specific labels are flipped matters more than the noise rate itself: some flip patterns happen to corrupt high-leverage early profiles while others leave them intact. At 20--30\% noise, accuracy degrades by $\sim$4pp but remains $+3$pp or more above the Vanilla RAG baseline (49.1\%), confirming that evidence profiles degrade gracefully rather than catastrophically under noisy outcome signals. The profile aggregation mechanism provides inherent robustness: even when noisy labels cause some spurious profiles to be written, the majority of accumulated signal remains correct, and the profile selection policy dilutes bad data as more queries are processed.

\section{Probe-Set Construction}
\label{app:probe_sensitivity}

The $N=73$ paired probe set combines the original 30 probes (selected with \texttt{MIN\_OVERLAP}~$\geq 15$, \texttt{MAX\_JACCARD}~$\leq 0.5$) with 43 additional probes selected at a relaxed \texttt{MIN\_OVERLAP}~$\geq 5$ from the same MuSiQue High-Reuse pool, 39 of which fall in the $[5,15)$ tier where verdict-noise is most pronounced. The relaxation broadens difficulty coverage without changing the conclusion: the per-subset Ours-RG lift on Haiku at $T=0$ is $\Delta=+0.070$ ($p=0.007$) on the original 30 and $\Delta=+0.123$ ($p=0.003$) on the new 43; for Sonnet the corresponding numbers are $+0.020$ ($p=0.35$, ceiling) and $+0.121$ ($p=0.001$). Treating the 73 probes as a single paired sample is therefore conservative; results stratified by subset are reported in the supplementary JSON summary.

\end{document}